\documentclass{article}

\usepackage{microtype}
\usepackage{graphicx}
\usepackage{booktabs} % for professional tables
\usepackage{graphicx}
\usepackage{multirow}
\usepackage{subcaption}
\usepackage{framed}
\usepackage{wrapfig}
\usepackage{natbib}
\usepackage{amsmath}
\usepackage{amssymb}
\usepackage{enumitem}
\usepackage{array}

\usepackage{hyperref}

\usepackage[accepted]{icml2018}

\usepackage{verbatim}

\newcommand\mypara[1]{\vspace{0mm}\noindent\textbf{#1}}

\icmltitlerunning{An Empirical Evaluation of Generic Convolutional and Recurrent Networks for Sequence Modeling}

\begin{document}

\twocolumn[
\icmltitle{An Empirical Evaluation of Generic Convolutional and Recurrent Networks for Sequence Modeling}

% It is OKAY to include author information, even for blind
% submissions: the style file will automatically remove it for you
% unless you've provided the [accepted] option to the icml2018
% package.

% List of affiliations: The first argument should be a (short)
% identifier you will use later to specify author affiliations
% Academic affiliations should list Department, University, City, Region, Country
% Industry affiliations should list Company, City, Region, Country

% You can specify symbols, otherwise they are numbered in order.
% Ideally, you should not use this facility. Affiliations will be numbered
% in order of appearance and this is the preferred way.
\icmlsetsymbol{equal}{*}

\begin{icmlauthorlist}
\icmlauthor{Shaojie Bai}{cmu-mld}
\icmlauthor{J. Zico Kolter}{cmu-csd}
\icmlauthor{Vladlen Koltun}{intel}
\end{icmlauthorlist}

\icmlaffiliation{cmu-mld}{Machine Learning Department, Carnegie Mellon University, Pittsburgh, PA, USA}
\icmlaffiliation{cmu-csd}{Computer Science Department, Carnegie Mellon University, Pittsburgh, PA, USA}
\icmlaffiliation{intel}{Intel Labs, Santa Clara, CA, USA}

\icmlcorrespondingauthor{Shaojie Bai}{shaojieb@cs.cmu.edu}
\icmlcorrespondingauthor{J. Zico Kolter}{zkolter@cs.cmu.edu}
\icmlcorrespondingauthor{Vladlen Koltun}{vkoltun@gmail.edu}

% You may provide any keywords that you
% find helpful for describing your paper; these are used to populate
% the "keywords" metadata in the PDF but will not be shown in the document
\icmlkeywords{Machine Learning, ICML, Deep Learning, Neural Networks, Sequence Modeling}

\vskip 0.3in
]

\printAffiliationsAndNotice{}

\makeatletter
\newcommand\footnoteref[1]{\protected@xdef\@thefnmark{\ref{#1}}\@footnotemark}
\makeatother

\newcommand{\charptbres}{1.31}
\newcommand{\chartextres}{1.45}
\newcommand{\wordptbres}{88.68}
\newcommand{\wordwikires}{45.19}
\newcommand{\wordlambadares}{1279}

\newcolumntype{P}[1]{>{\centering\arraybackslash}p{#1}}

% this must go after the closing bracket ] following \twocolumn[ ...

% This command actually creates the footnote in the first column
% listing the affiliations and the copyright notice.
% The command takes one argument, which is text to display at the start of the footnote.
% The \icmlEqualContribution command is standard text for equal contribution.
% Remove it (just {}) if you do not need this facility.

 % leave blank if no need to mention equal contribution
%\printAffiliationsAndNotice{\icmlEqualContribution} % otherwise use the standard text.

\begin{abstract}
For most deep learning practitioners, sequence modeling is synonymous with recurrent networks. Yet recent results indicate that convolutional architectures can outperform recurrent networks on tasks such as audio synthesis and machine translation. Given a new sequence modeling task or dataset, which architecture should one use? We conduct a systematic evaluation of generic convolutional and recurrent architectures for sequence modeling. The models are evaluated across a broad range of standard tasks that are commonly used to benchmark recurrent networks. Our results indicate that a simple convolutional architecture outperforms canonical recurrent networks such as LSTMs across a diverse range of tasks and datasets, while demonstrating longer effective memory. We conclude that the common association between sequence modeling and recurrent networks should be reconsidered, and convolutional networks should be regarded as a natural starting point for sequence modeling tasks. To assist related work, we have made code available at \url{http://github.com/locuslab/TCN}.
\end{abstract}

\section{Introduction}

Deep learning practitioners commonly regard recurrent architectures as the default starting point for sequence modeling tasks. The sequence modeling chapter in the canonical textbook on deep learning is titled ``Sequence Modeling: Recurrent and Recursive Nets''~\cite{Goodfellow-et-al-2016}, capturing the common association of sequence modeling and recurrent architectures. A well-regarded recent online course on ``Sequence Models'' focuses exclusively on recurrent architectures~\cite{Ng2018}.

On the other hand, recent research indicates that certain convolutional architectures can reach state-of-the-art accuracy in audio synthesis, word-level language modeling, and machine translation~\cite{waveNet,kalchbrenner2016neural,dauphinGatedConv,Gehring2017:ACL,gehring2017convolutional}. This raises the question of whether these successes of convolutional sequence modeling are confined to specific application domains or whether a broader reconsideration of the association between sequence processing and recurrent networks is in order.

We address this question by conducting a systematic empirical evaluation of convolutional and recurrent architectures on a broad range of sequence modeling tasks. We specifically target a comprehensive set of tasks that have been repeatedly used to compare the effectiveness of different recurrent network architectures. These tasks include polyphonic music modeling, word- and character-level language modeling, as well as synthetic stress tests that had been deliberately designed and frequently used to benchmark RNNs. Our evaluation is thus set up to compare convolutional and recurrent approaches to sequence modeling on the recurrent networks' ``home turf''.

To represent convolutional networks, we describe a generic temporal convolutional network (TCN) architecture that is applied across all tasks. This architecture is informed by recent research, but is deliberately kept simple, combining some of the best practices of modern convolutional architectures. It is compared to canonical recurrent architectures such as LSTMs and GRUs.

The results suggest that TCNs convincingly outperform baseline recurrent architectures across a broad range of sequence modeling tasks. This is particularly notable because the tasks include diverse benchmarks that have commonly been used to evaluate recurrent network designs~\cite{chung2014empirical,Pascanu2014:ICLR,jozefowicz2015empirical,Zhang2016NIPS}. This indicates that the recent successes of convolutional architectures in applications such as audio processing are not confined to these domains.

To further understand these results, we analyze more deeply the memory retention characteristics of recurrent networks. We show that despite the theoretical ability of recurrent architectures to capture infinitely long history, TCNs exhibit substantially longer memory, and are thus more suitable for domains where a long history is required.

To our knowledge, the presented study is the most extensive systematic comparison of convolutional and recurrent architectures on sequence modeling tasks. The results suggest that the common association between sequence modeling and recurrent networks should be reconsidered. The TCN architecture appears not only more accurate than canonical recurrent networks such as LSTMs and GRUs, but also simpler and clearer. It may therefore be a more appropriate starting point in the application of deep networks to sequences.

\section{Background}
\label{sec:background}

Convolutional networks~\cite{LeCun1989} have been applied to sequences for decades~\cite{sejnowski,hintonConnectionist}. They were used prominently for speech recognition in the 80s and 90s~\cite{waibel,bottou}. ConvNets were subsequently applied to NLP tasks such as part-of-speech tagging and semantic role labelling~\cite{collobertUnified,Collobert2011,Santos2014:ICML}. More recently, convolutional networks were applied to sentence classification~\cite{kalchbrennerGB14,kim2014convolutional} and document classification~\cite{Zhang2015:NIPS,Conneau2017,Johnson2015,Johnson2017}. Particularly inspiring for our work are the recent applications of convolutional architectures to machine translation~\cite{kalchbrenner2016neural,Gehring2017:ACL,gehring2017convolutional}, audio synthesis~\cite{waveNet}, and language modeling~\cite{dauphinGatedConv}.

Recurrent networks are dedicated sequence models that maintain a vector of hidden activations that are propagated through time~\cite{Elman90findstructure,Werbos1990,Graves2012}. This family of architectures has gained tremendous popularity due to prominent applications to language modeling~\cite{Sutskever2011,graves2013generating,Hermans2013} and machine translation~\cite{sutskeverSeqToSeq,Bahdanau2015}. The intuitive appeal of recurrent modeling is that the hidden state can act as a representation of everything that has been seen so far in the sequence. Basic RNN architectures are notoriously difficult to train~\citep{bengio1994learning,pascanu2013difficulty} and more elaborate architectures are commonly used instead, such as the LSTM~\cite{hochreiterLSTM} and the GRU~\cite{choGRU}. Many other architectural innovations and training techniques for recurrent networks have been introduced and continue to be actively explored~\cite{HihiBengio1995,Schuster1997,gers2002learning,koutnikClockwork,leIRNN,Ba2016layer,wuMIRNN,kruegerZoneout,merityRegOpt,campos2018skip}.

Multiple empirical studies have been conducted to evaluate the effectiveness of different recurrent architectures. These studies have been motivated in part by the many degrees of freedom in the design of such architectures. \citet{chung2014empirical} compared different types of recurrent units (LSTM vs.\ GRU) on the task of polyphonic music modeling. \citet{Pascanu2014:ICLR} explored different ways to construct deep RNNs and evaluated the performance of different architectures on polyphonic music modeling, character-level language modeling, and word-level language modeling. \citet{jozefowicz2015empirical} searched through more than ten thousand different RNN architectures and evaluated their performance on various tasks. They concluded that if there were ``architectures much better than the LSTM'', then they were ``not trivial to find''. \citet{greffOdyssey} benchmarked the performance of eight LSTM variants on speech recognition, handwriting recognition, and polyphonic music modeling. They also found that ``none of the variants can improve upon the standard LSTM architecture significantly''. \citet{Zhang2016NIPS} systematically analyzed the connecting architectures of RNNs and evaluated different architectures on character-level language modeling and on synthetic stress tests. \citet{Melis2018} benchmarked LSTM-based architectures on word-level and character-level language modeling, and concluded that ``LSTMs outperform the more recent models''.

Other recent works have aimed to combine aspects of RNN and CNN architectures.  This includes the Convolutional LSTM \cite{xingjian2015convolutional}, which replaces the fully-connected layers in an LSTM with convolutional layers to allow for additional structure in the recurrent layers; the Quasi-RNN model \cite{bradbury2016quasi} that interleaves convolutional layers with simple recurrent layers; and the dilated RNN \cite{chang2017dilated}, which adds dilations to recurrent architectures.  While these combinations show promise in combining the desirable aspects of both types of architectures, our study here focuses on a comparison of generic convolutional and recurrent architectures.

While there have been multiple thorough evaluations of RNN architectures on representative sequence modeling tasks, we are not aware of a similarly thorough comparison of convolutional and recurrent approaches to sequence modeling. (\citet{Yin2017} have reported a comparison of convolutional and recurrent networks for sentence-level and document-level classification tasks. In contrast, sequence modeling calls for architectures that can synthesize whole sequences, element by element.) Such comparison is particularly intriguing in light of the aforementioned recent success of convolutional architectures in this domain. Our work aims to compare generic convolutional and recurrent architectures on typical sequence modeling tasks that are commonly used to benchmark RNN variants themselves~\cite{Hermans2013,leIRNN,jozefowicz2015empirical,Zhang2016NIPS}.

\section{Temporal Convolutional Networks}
\label{technical-sec}

We begin by describing a generic architecture for convolutional sequence prediction.
Our aim is to distill the best practices in convolutional network design into a simple
architecture that can serve as a convenient but powerful starting point. We refer to the
presented architecture as a temporal convolutional network (TCN), emphasizing that we
adopt this term not as a label for a truly new architecture, but as a simple descriptive
term for a family of architectures. (Note that the term has been used before~\cite{Lea2017}.)
The distinguishing characteristics of TCNs are: 1) the convolutions in the architecture are
causal, meaning that there is no information ``leakage'' from future to past; 2) the
architecture can take a sequence of any length and map it to an output sequence of the
same length, just as with an RNN. Beyond this, we emphasize how to build very long effective
history sizes (i.e., the ability for the networks to look very far into the past to make a
prediction) using a combination of very deep networks (augmented with residual layers) and
dilated convolutions.

Our architecture is informed by recent convolutional architectures for sequential
data~\cite{waveNet,kalchbrenner2016neural,dauphinGatedConv,Gehring2017:ACL,gehring2017convolutional},
but is distinct from all of them and was designed from first principles to combine simplicity,
autoregressive prediction, and very long memory. For example, the TCN is much simpler than
WaveNet~\cite{waveNet} (no skip connections across layers, conditioning, context stacking,
or gated activations).

Compared to the language modeling architecture of \citet{dauphinGatedConv}, TCNs do not use
gating mechanisms and have much longer memory.

\subsection{Sequence Modeling}

Before defining the network structure, we highlight the nature of the sequence
modeling task.  Suppose that we are given an input sequence $x_0,\ldots,x_T$,
and wish to predict some corresponding outputs $y_0,\ldots,y_T$ at each time.
The key constraint is that to predict the output $y_t$ for some time $t$, we are
constrained to only use those inputs that have been previously observed:
$x_0,\ldots,x_t$.  Formally, a sequence modeling network is any function ${f :
\mathcal{X}^{T+1} \rightarrow \mathcal{Y}^{T+1}}$ that produces the mapping
\begin{equation}
\hat{y}_0,\ldots,\hat{y}_T = f(x_0, \ldots, x_T)
\end{equation}
if it satisfies the causal constraint that $y_t$ depends only on
${x_0,\ldots,x_t}$ and not on any ``future'' inputs ${x_{t+1},\ldots,x_T}$. The
goal of learning in the sequence modeling setting is to find a network $f$
that minimizes some expected loss between the actual outputs and the predictions,
$L(y_0,\ldots,y_T, f(x_0,\ldots,x_T))$, where the sequences and outputs are drawn
according to some distribution.

This formalism encompasses many settings such as auto-regressive prediction
(where we try to predict some signal given its past) by setting the target
output to be simply the input shifted by one time step. It does not,
however, directly capture domains such as machine translation, or
sequence-to-sequence prediction in general, since in these cases the entire
input sequence (including ``future" states) can be used to predict each
output (though the techniques can naturally be extended to work in such
settings).

\begin{figure*}
\centering
\begin{tabular}{@{}ccc@{}}
  \includegraphics[height=4.3cm]{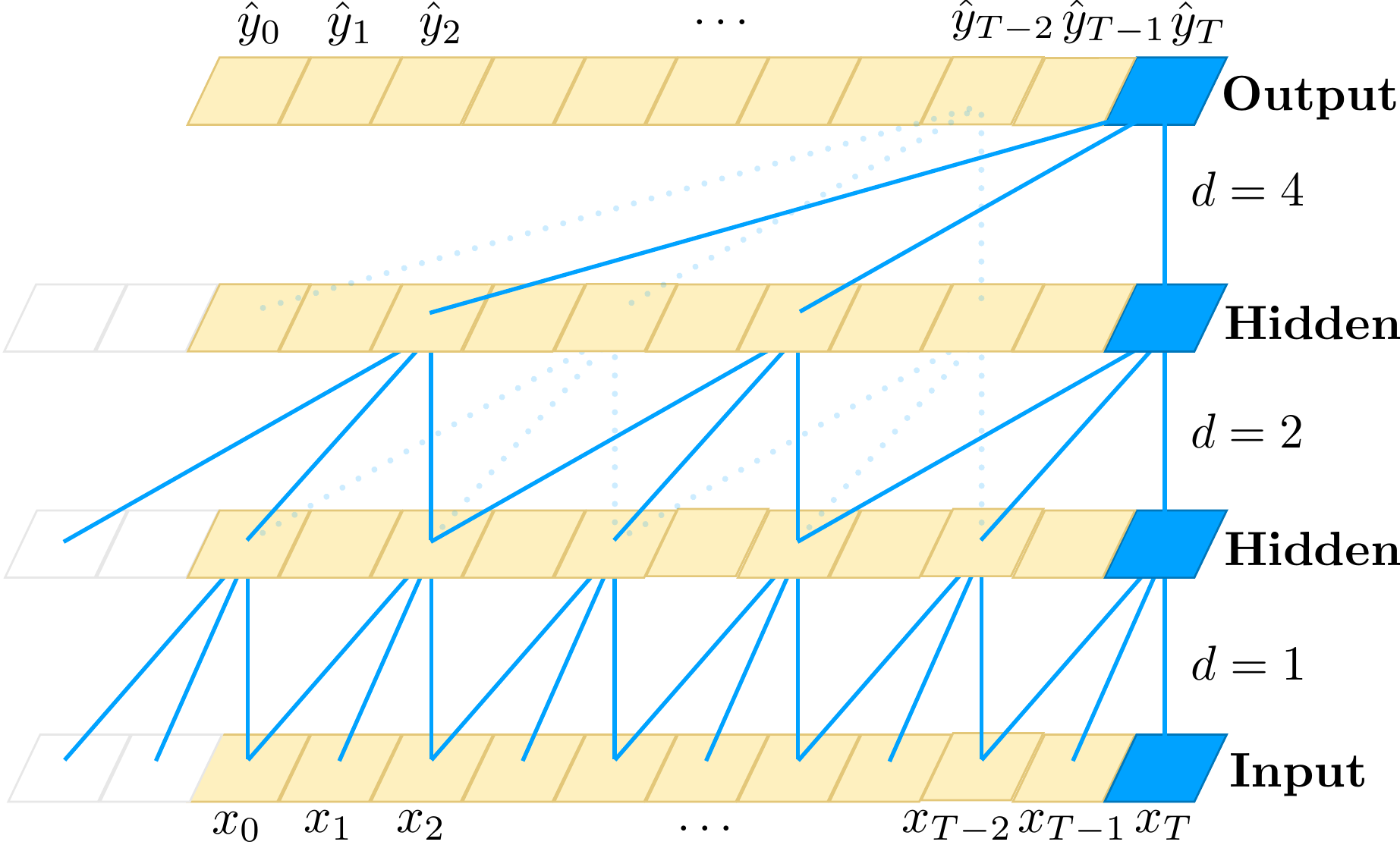} & \includegraphics[height=4.3cm]{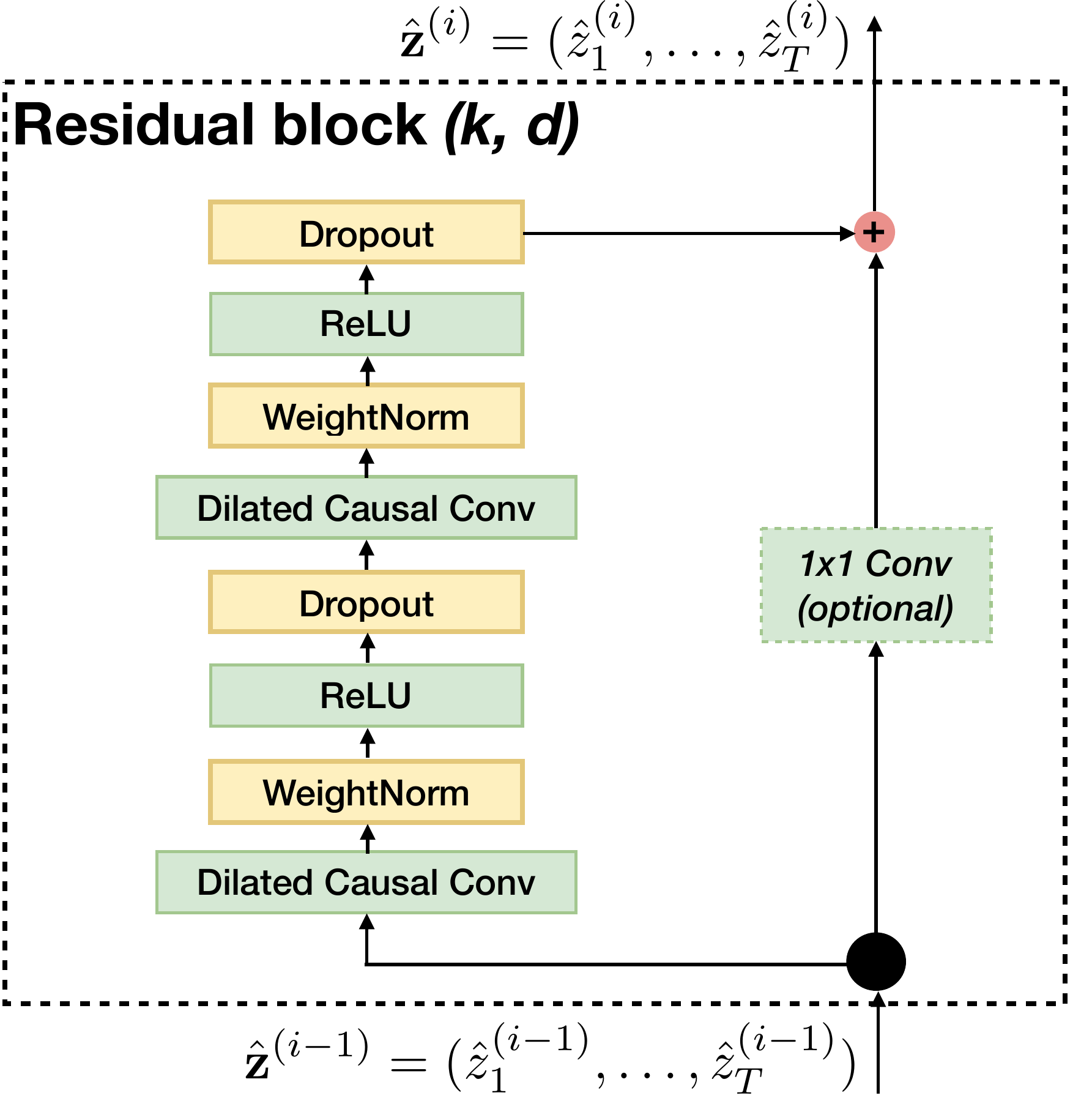} & \includegraphics[height=4.3cm]{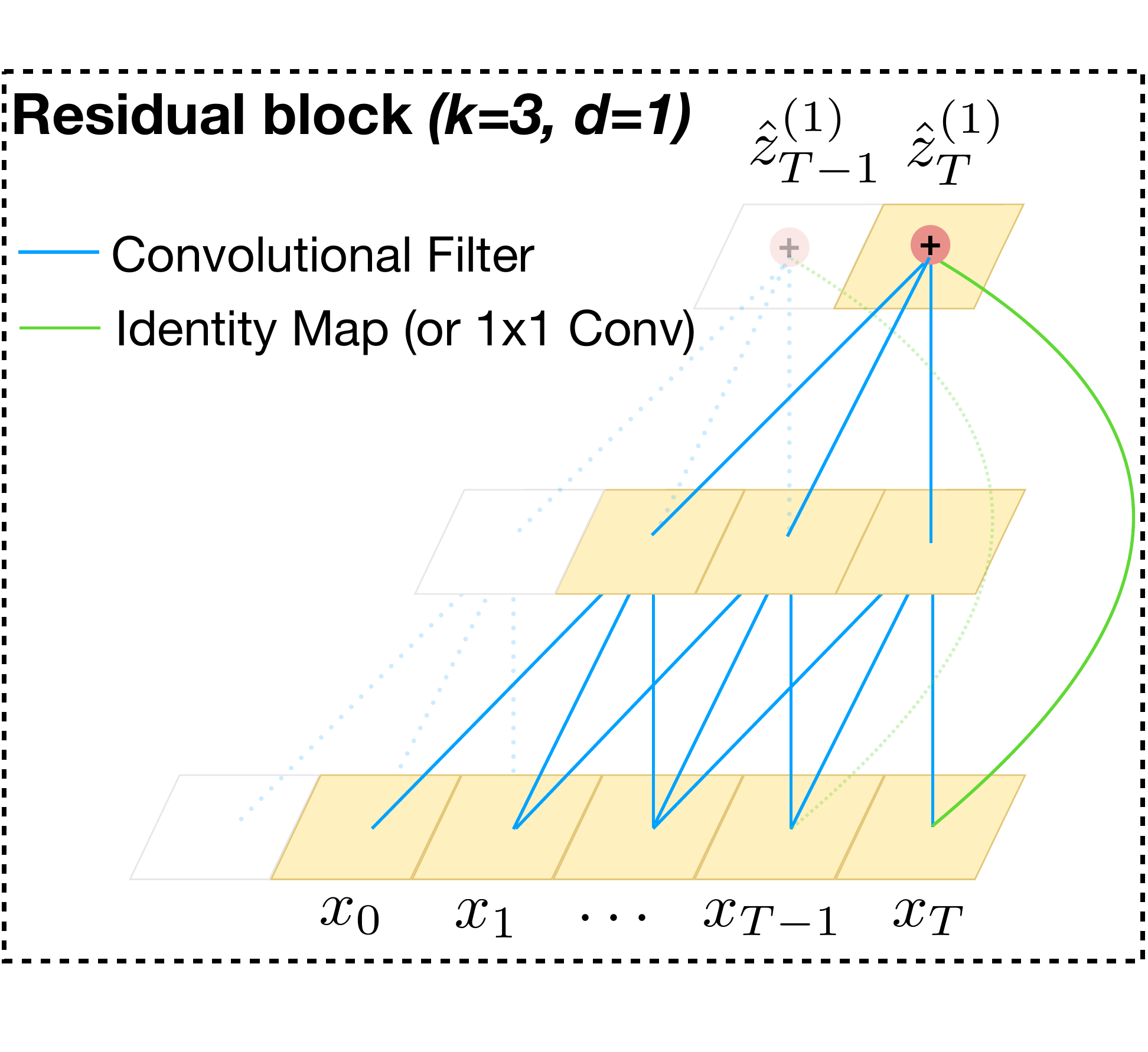} \\
  \small (a) & \small (b) & \small (c)
\end{tabular}
\vspace{-.1in}
\caption{Architectural elements in a TCN. (a) A dilated causal convolution with dilation factors $d=1,2,4$ and filter size $k=3$. The receptive field is able to cover all values from the input sequence. (b) TCN residual block. An 1x1 convolution is added
when residual input and output have different dimensions. (c) An example of residual connection in a TCN. The blue lines are filters in the residual function, and the green lines are identity mappings.}
\label{fig:TCN}
\vspace{-.15in}
\end{figure*}

\subsection{Causal Convolutions}

As mentioned above, the TCN is based upon two principles: the fact that the
network produces an output of the same length as the input, and the fact that
there can be no leakage from the future into the past.  To accomplish the
first point, the TCN uses a 1D fully-convolutional network (FCN) architecture
\citep{long2015fully}, where each hidden layer is the same length as the input
layer, and zero padding of length $(\mbox{kernel size}-1)$ is added to keep
subsequent layers the same length as previous ones.   To achieve the second
point, the TCN uses \emph{causal convolutions}, convolutions where an
output at time $t$ is convolved only with elements from time $t$ and earlier in the
previous layer.

To put it simply: TCN $=$ 1D FCN $+$ causal convolutions.

Note that this is essentially the same architecture as the time delay neural network proposed
nearly 30 years ago by \citet{waibel}, with the sole tweak of zero padding to ensure equal
sizes of all layers.

A major disadvantage of this basic design is that in order to
achieve a long effective history size, we need an extremely
deep network or very large filters, neither of which were particularly feasible
when the methods were first introduced.  Thus, in the following sections, we
describe how techniques from modern convolutional architectures can be integrated
into a TCN to allow for both very deep networks and very long effective history.

\subsection{Dilated Convolutions}

A simple causal convolution is only able to look back at a history with size linear
in the depth of the network. This makes it challenging to apply the aforementioned causal
convolution on sequence tasks, especially those requiring longer history.
Our solution here, following the work of~\citet{waveNet}, is to employ dilated convolutions
that enable an exponentially large receptive field~\citep{dilatedConv}. More formally, for
a 1-D sequence input $\mathbf{x} \in \mathbb{R}^{n}$ and a filter $f: \{0, \dots, k-1\}
\rightarrow \mathbb{R}$, the dilated convolution operation $F$ on element $s$ of
the sequence is defined as
\vspace{-2mm}
\begin{equation}
F(s) = (\mathbf{x} *_d f)(s) = \sum_{i=0}^{k-1} f(i) \cdot \mathbf{x}_{s-d \cdot i}
\end{equation}
where $d$ is the dilation factor, $k$ is the filter size, and $s-d \cdot i$ accounts
for the direction of the past. Dilation is thus equivalent to introducing a fixed
step between every two adjacent filter taps. When $d=1$, a dilated
convolution reduces to a regular convolution. Using larger dilation enables
an output at the top level to represent a wider range of inputs, thus
effectively expanding the receptive field of a ConvNet.

This gives us two ways to increase the receptive field of the TCN: choosing
larger filter sizes $k$ and increasing the dilation factor $d$, where the
effective history of one such layer is $(k-1)d$.  As is common when using dilated
convolutions, we increase $d$ exponentially with the depth of the
network (i.e., $d = O(2^i)$ at level $i$ of the network). This ensures
that there is some filter that hits each input within the effective history, while
also allowing for an extremely large effective history using deep networks.  We
provide an illustration in Figure \ref{fig:TCN}(a).

\subsection{Residual Connections}

A residual block~\cite{he2016deep} contains a branch leading out to a series of
transformations $\mathcal{F}$, whose outputs are added to the input $\mathbf{x}$
of the block:
\begin{equation}
o = \text{Activation}(\mathbf{x} + \mathcal{F}(\mathbf{x}))
\end{equation}
This effectively allows layers to learn modifications to the identity
mapping rather than the entire transformation, which has repeatedly been shown
to benefit very deep networks.

Since a TCN's receptive field depends on the network depth $n$ as well
as filter size $k$ and dilation factor $d$, stabilization of deeper and larger TCNs
becomes important. For example, in a case where the prediction could depend on
a history of size $2^{12}$ and a high-dimensional input sequence, a network of
up to 12 layers could be needed. Each layer, more specifically, consists of
multiple filters for feature extraction. In our design of the generic TCN model,
we therefore employ a generic residual module in place of a convolutional layer.

The residual block for our baseline TCN is shown in Figure~\ref{fig:TCN}(b).
Within a residual block, the TCN has two layers of dilated causal convolution
and non-linearity, for which we used the rectified linear unit (ReLU)
\citep{nair2010rectified}. For normalization, we applied
weight normalization \citep{Salimans2016} to the convolutional filters.
In addition, a spatial dropout \citep{srivastava2014dropout} was added after each
dilated convolution for regularization: at each training step, a whole channel is
zeroed out.

However, whereas in standard ResNet the input is added directly
to the output of the residual function, in TCN (and ConvNets in general) the
input and output could have different widths. To account for discrepant input-output
widths, we use an additional 1x1 convolution to ensure
that element-wise addition~$\oplus$ receives tensors of the same shape (see Figure
\ref{fig:TCN}(b,c)).

\subsection{Discussion}

We conclude this section by listing several advantages and disadvantages of using
TCNs for sequence modeling.

\vspace{-.1in}
\begin{itemize}[leftmargin=*]
\setlength{\itemsep}{0mm}
\setlength{\parskip}{1mm}{}
\item \textbf{Parallelism}. Unlike in RNNs where the predictions for later timesteps
must wait for their predecessors to complete, convolutions can be done in parallel
since the same filter is used in each layer. Therefore, in both training and evaluation,
a long input sequence can be processed as a whole in TCN, instead of sequentially as in RNN.

\item \textbf{Flexible receptive field size}. A TCN can change
its receptive field size in multiple ways. For instance, stacking more
dilated (causal) convolutional layers, using larger dilation factors, or
increasing the filter size are all viable options (with possibly different
interpretations). TCNs thus afford better control of the model's memory size,
and are easy to adapt to different domains.

\item \textbf{Stable gradients}. Unlike recurrent architectures, TCN has
a backpropagation path different from the temporal direction of the
sequence. TCN thus avoids the problem of exploding/vanishing
gradients, which is a major issue for RNNs (and which led to the
development of LSTM, GRU, HF-RNN~\citep{MartensRNNHF}, etc.).

\item \textbf{Low memory requirement for training}. Especially in
the case of a long input sequence, LSTMs and GRUs can easily use up a lot
of memory to store the partial results for their multiple cell gates. However,
in a TCN the filters are shared across a layer, with the backpropagation path depending only
on network depth. Therefore in practice, we found gated RNNs likely to use up to a
multiplicative factor more memory than TCNs.

\item \textbf{Variable length inputs}. Just like RNNs, which model inputs with
variable lengths in a recurrent way, TCNs can also take in inputs of arbitrary
lengths by sliding the 1D convolutional kernels. This means that TCNs can be adopted as
drop-in replacements for RNNs for sequential data of arbitrary length.
\end{itemize}

There are also two notable disadvantages to using TCNs.
\begin{itemize}[leftmargin=*]
\setlength{\itemsep}{0mm}
\setlength{\parskip}{1mm}
\item \textbf{Data storage during evaluation}. In evaluation/testing, RNNs only
need to maintain a hidden state and take in a current input $x_t$ in order to
generate a prediction. In other words, a ``summary'' of the entire history is
provided by the fixed-length set of vectors $h_t$, and the actual
observed sequence can be discarded.  In contrast, TCNs need to take in the raw
sequence up to the effective history length, thus
possibly requiring more memory during evaluation.

\item \textbf{Potential parameter change for a transfer of domain}. Different
domains can have different requirements on the amount of history the model needs
in order to predict. Therefore, when transferring a model from a domain where only
little memory is needed (i.e., small $k$ and $d$) to a domain where much longer
memory is required (i.e., much larger $k$ and $d$), TCN may perform poorly for
not having a sufficiently large receptive field.
\end{itemize}

\section{Sequence Modeling Tasks}
\label{app-tasks}

We evaluate TCNs and RNNs on tasks that have been commonly used to benchmark
the performance of different RNN sequence modeling architectures~\cite{Hermans2013,
chung2014empirical,Pascanu2014:ICLR,leIRNN,jozefowicz2015empirical,Zhang2016NIPS}.
The intention is to conduct the evaluation on the ``home turf'' of RNN sequence
models. We use a comprehensive set of synthetic stress tests along with real-world datasets from multiple domains.

\mypara{The adding problem.} In this task, each input consists of a length-$n$
sequence of depth 2, with all values randomly chosen in $[0, 1]$, and the
second dimension being all zeros except for two elements that are marked by 1.
The objective is to sum the two random values whose second dimensions are
marked by 1. Simply predicting the sum to be 1 should give an MSE of about
0.1767. First introduced by \citet{hochreiterLSTM}, the adding problem has been used repeatedly as a stress test for sequence models~\cite{MartensRNNHF,pascanu2013difficulty,leIRNN,arjovsky2016unitary,Zhang2016NIPS}.

\mypara{Sequential MNIST and P-MNIST.} Sequential MNIST is frequently used to
test a recurrent network's ability to retain information from the distant past~\citep{leIRNN,Zhang2016NIPS,wisdom2016full,cooijmans2016recurrent,kruegerZoneout,pmlr-v70-jing17a}.
In this task, MNIST images~\citep{Lecun98gradient} are presented to the
model as a $784 \!\times\! 1$ sequence for digit classification. In the more
challenging P-MNIST setting, the order of the sequence is permuted at random~\cite{leIRNN,arjovsky2016unitary,wisdom2016full,kruegerZoneout}.

\mypara{Copy memory.} In this task, each input sequence has length
$T+20$. The first 10 values are chosen randomly among the digits ${1,\ldots,8}$, with the rest
being all zeros, except for the last 11 entries that are filled with the digit `9' (the
first `9' is a delimiter). The goal is to generate an output of
the same length that is zero everywhere except the last 10 values after the
delimiter, where the model is expected to repeat the 10 values it encountered at the
start of the input. This task was used in prior works such as
\citet{Zhang2016NIPS,arjovsky2016unitary,wisdom2016full,pmlr-v70-jing17a}.

\begin{table*}[t]
\def\arraystretch{1.3}
\small
\centering
\caption{Evaluation of TCNs and recurrent architectures on synthetic stress tests, polyphonic music modeling, character-level language modeling, and word-level language modeling. The generic TCN architecture outperforms canonical recurrent networks across a comprehensive suite of tasks and datasets. Current state-of-the-art results are listed in the supplement. $\ {}^h$~means that higher is better. ${}^\ell$~means that lower is better.}
\vspace{2mm}
{\begin{tabular}{l*{6}{c}}
\toprule
\multirow{2}{*}{Sequence Modeling Task} & \multirow{2}{*}{Model Size ($\approx$)} & \multicolumn{4}{c}{Models} \\
\cline{3-6}
\textbf    & & LSTM & GRU & RNN & \textbf{TCN} \\
\midrule
Seq. MNIST (accuracy${}^h$)             & 70K   & 87.2    & 96.2              & 21.5   & \textbf{99.0}   \\
Permuted MNIST (accuracy)               & 70K   & 85.7    & 87.3              & 25.3   & \textbf{97.2}   \\
Adding problem $T$=600 (loss${}^\ell$)  & 70K   & 0.164   & \textbf{5.3e-5}   & 0.177  &  \textbf{5.8e-5}  \\
Copy memory $T$=1000 (loss)             & 16K   & 0.0204  & 0.0197            & 0.0202 & \textbf{3.5e-5} \\
Music JSB Chorales (loss)               & 300K   & 8.45           & 8.43     & 8.91    & \textbf{8.10}  \\
Music Nottingham (loss)                 & 1M     & 3.29           & 3.46     & 4.05    & \textbf{3.07} \\
Word-level PTB (perplexity${}^\ell$)    & 13M    & \textbf{78.93} & 92.48    & 114.50  & \wordptbres  \\
Word-level Wiki-103 (perplexity)        & -      & 48.4           & -        & -       & \textbf{\wordwikires} \\
Word-level LAMBADA (perplexity)         &  -     & 4186           & -        & 14725   & \textbf{\wordlambadares}  \\
Char-level PTB (bpc${}^\ell$)           & 3M     & 1.36           & 1.37     & 1.48    & \textbf{\charptbres} \\
Char-level text8 (bpc)                  & 5M     & 1.50           & 1.53     & 1.69    & \textbf{\chartextres}  \\
\bottomrule
\end{tabular}}
\label{bigtable}
\end{table*}

\mypara{JSB Chorales and Nottingham.} JSB Chorales~\citep{allan2005harmonising} is a
polyphonic music dataset consisting of the entire corpus of 382 four-part
harmonized chorales by J. S. Bach. Each input is a sequence of elements. Each element is an 88-bit binary code that corresponds to the 88 keys on a
piano, with 1 indicating a key that is pressed at a given time.
Nottingham is a polyphonic music dataset based on a collection of 1,200 British and American folk tunes, and is much larger than JSB Chorales. JSB Chorales and Nottingham have been used in numerous empirical investigations of recurrent sequence modeling~\citep{chung2014empirical,Pascanu2014:ICLR,jozefowicz2015empirical,greffOdyssey}. The performance on both tasks is measured in terms of negative log-likelihood (NLL).

\mypara{PennTreebank.} We used the PennTreebank (PTB)~\citep{Marcus93buildinga} for both
character-level and word-level language modeling. When used as a character-level language corpus,
PTB contains 5,059K characters for training, 396K for validation, and 446K for testing, with an
alphabet size of 50. When used as a word-level language corpus, PTB contains
888K words for training, 70K for validation, and 79K for testing, with a vocabulary size of 10K.
This is a highly studied but relatively small language modeling
dataset~\citep{miyamoto2016gated,kruegerZoneout,merityRegOpt}.

\mypara{Wikitext-103.} Wikitext-103 \citep{merity2016pointer} is almost 110
times as large as PTB, featuring a vocabulary size of about 268K.  The dataset
contains 28K Wikipedia articles (about 103 million words) for training, 60
articles (about 218K words) for validation, and 60 articles (246K words) for
testing. This is a more representative and realistic dataset than PTB, with a much
larger vocabulary that includes many rare words, and has been used in
\citet{merity2016pointer, grave2016improving, dauphinGatedConv}.

\mypara{LAMBADA.} Introduced by \citet{paperno2016lambada}, LAMBADA is a dataset
comprising 10K passages extracted from novels, with an average of 4.6 sentences
as context, and 1 target sentence the last word of which is to be predicted. This
dataset was built so that a person can easily guess the missing word when given the
context sentences, but not when given only the target sentence without the context sentences. Most of the
existing models fail on LAMBADA~\cite{paperno2016lambada, grave2016improving}. In general, better results on LAMBADA
indicate that a model is better at capturing information from longer and broader
context. The training data for LAMBADA is the full text of 2,662 novels with more
than 200M words. The vocabulary size is about 93K.

\mypara{text8.} We also used the text8 dataset for character-level language
modeling \citep{mikolov2012subword}. text8 is about 20 times
larger than PTB, with about 100M characters from Wikipedia (90M for training, 5M
for validation, and 5M for testing). The corpus contains 27 unique alphabets.

\section{Experiments}
\label{experiments-sec}

We compare the generic TCN architecture described in Section~\ref{technical-sec}
to canonical recurrent architectures, namely LSTM, GRU, and vanilla RNN, with
standard regularizations. All experiments reported in this section used exactly
the same TCN architecture, just varying the depth of the network $n$ and occasionally
the kernel size $k$ so that the receptive field covers enough context for
predictions. We use an exponential dilation $d=2^i$ for layer $i$ in the network,
and the Adam optimizer \citep{kingma2014adam} with learning rate $0.002$ for TCN, unless
otherwise noted. We also empirically find that gradient clipping
helped convergence, and we pick the maximum norm for clipping from
$[0.3, 1]$. When training recurrent models, we use grid search to find a
good set of hyperparameters (in particular, optimizer, recurrent drop
$p \in [0.05, 0.5]$, learning rate, gradient clipping, and initial forget-gate
bias), while keeping the network around the same size as TCN. No other
architectural elaborations, such as gating mechanisms or skip connections,
were added to either TCNs or RNNs. Additional details and controlled experiments are
provided in the supplementary material.

\subsection{Synopsis of Results}

A synopsis of the results is shown in Table~\ref{bigtable}. Note that on several of
these tasks, the generic, canonical recurrent architectures we study (e.g., LSTM, GRU) are
not the state-of-the-art. (See the supplement for more details.) With this caveat, the results strongly
suggest that the generic TCN architecture \emph{with minimal tuning} outperforms canonical
recurrent architectures across a broad variety of sequence modeling tasks that are commonly
used to benchmark the performance of recurrent architectures themselves.
We now analyze these results in more detail.

\subsection{Synthetic Stress Tests}
\label{baseline-tasks}

\begin{figure}[t]
    \centering
    \begin{subfigure}[t]{0.23\textwidth}
        \centering
        \includegraphics[width=\textwidth]{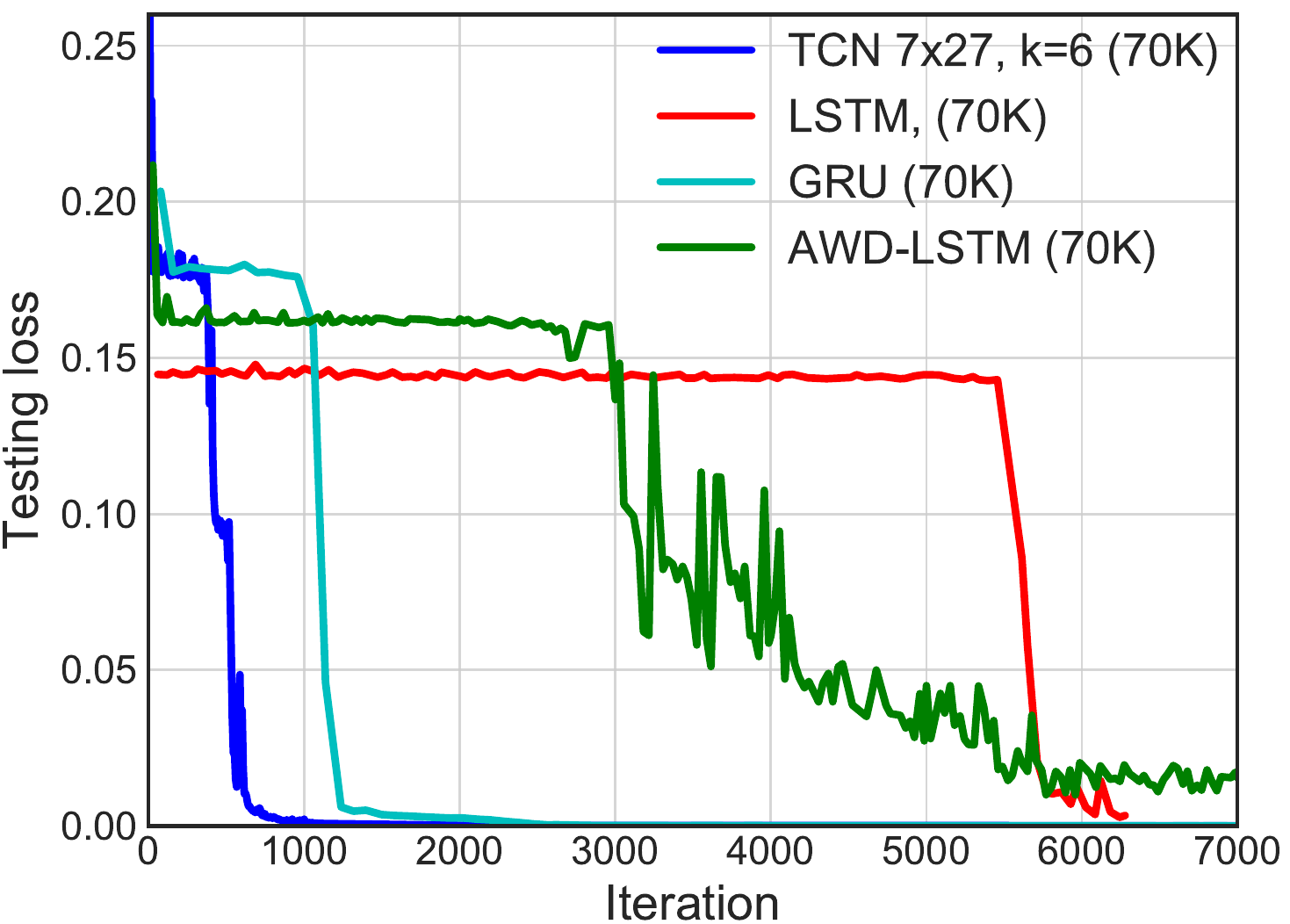}
        \caption{$T=200$ \label{adding200-figure}}
    \end{subfigure}
    ~
    \begin{subfigure}[t]{0.23\textwidth}
        \centering
        \includegraphics[width=\textwidth]{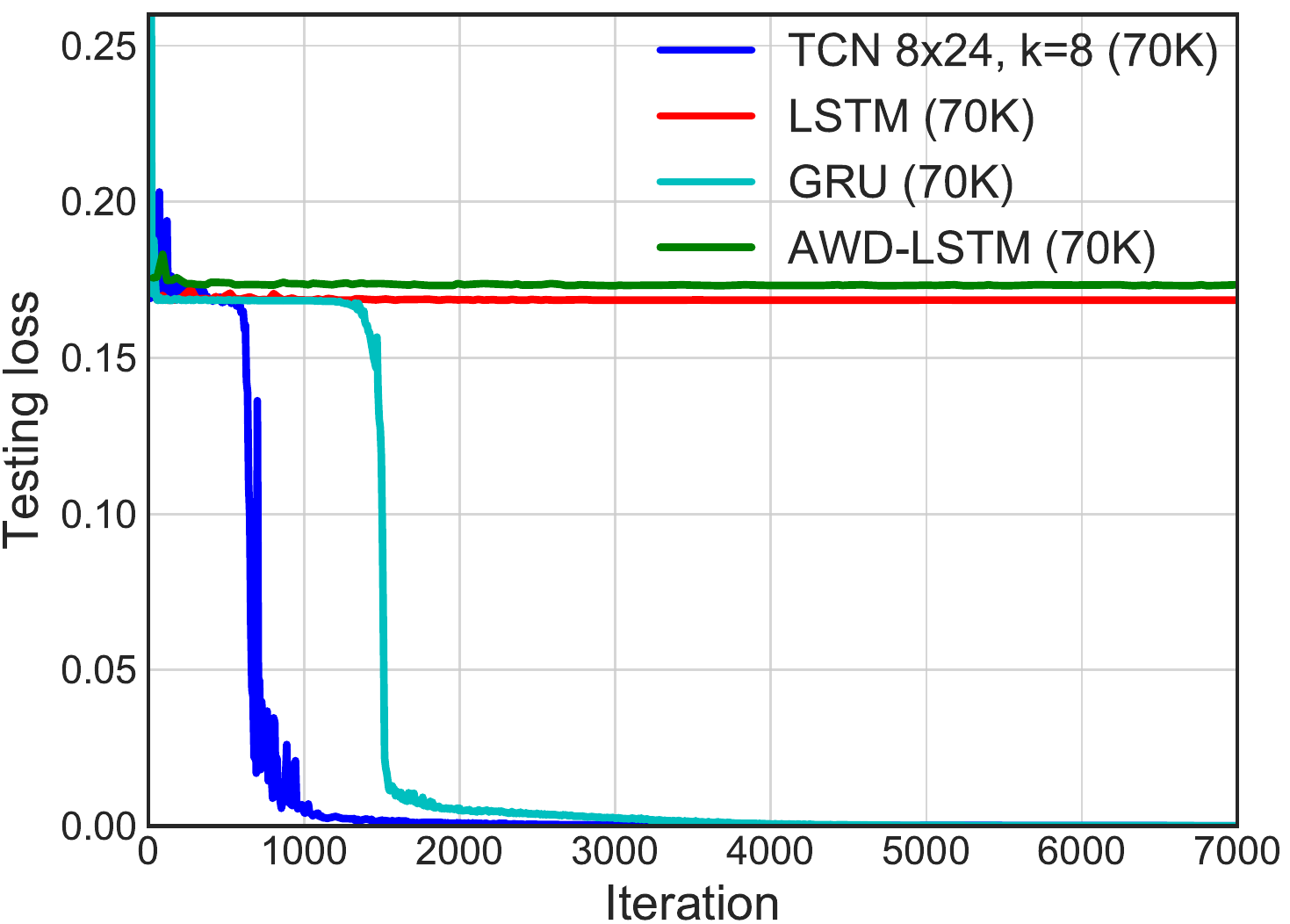}
        \caption{$T=600$ \label{adding600-figure}}
    \end{subfigure}
    \vspace{-.05in}
    \caption{Results on the adding problem for different sequence lengths $T$. TCNs outperform recurrent architectures.}
    \label{adding-figure}
\vspace{-1mm}
\end{figure}

\mypara{The adding problem.}
Convergence results for the adding problem, for problem sizes $T=200$ and $600$,
are shown in Figure \ref{adding-figure}. All models were chosen to have roughly
70K parameters.  TCNs quickly converged to a virtually
perfect solution (i.e., MSE near 0). GRUs also performed quite
well, albeit slower to converge than TCNs. LSTMs and vanilla RNNs
performed significantly worse.

\begin{figure}[t]
    \centering
    \begin{subfigure}[t]{0.23\textwidth}
        \centering
        \includegraphics[width=\textwidth]{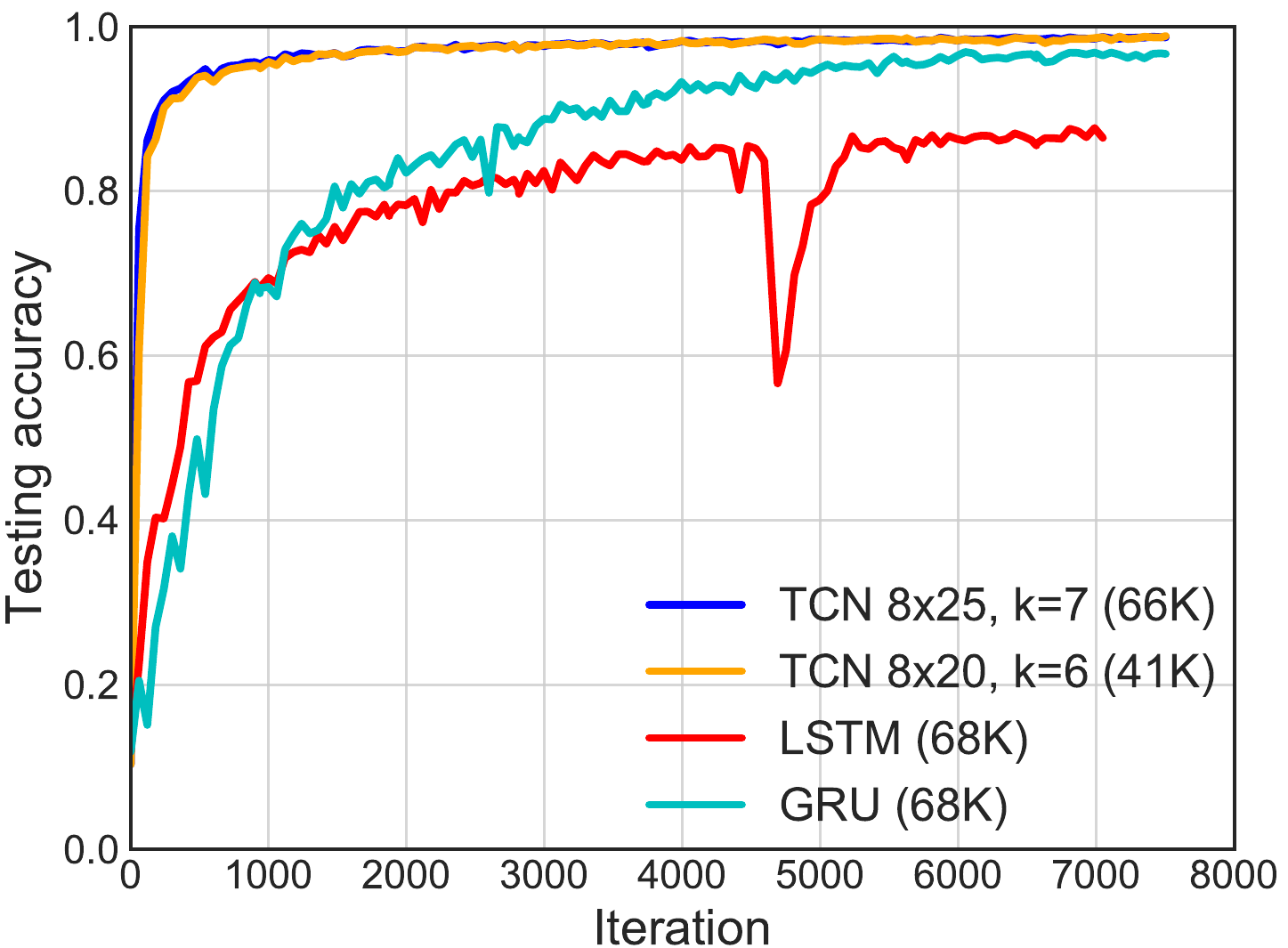}
        \caption{Sequential MNIST \label{MNIST-figure}}
    \end{subfigure}
    ~
    \begin{subfigure}[t]{0.23\textwidth}
        \centering
        \includegraphics[width=\textwidth]{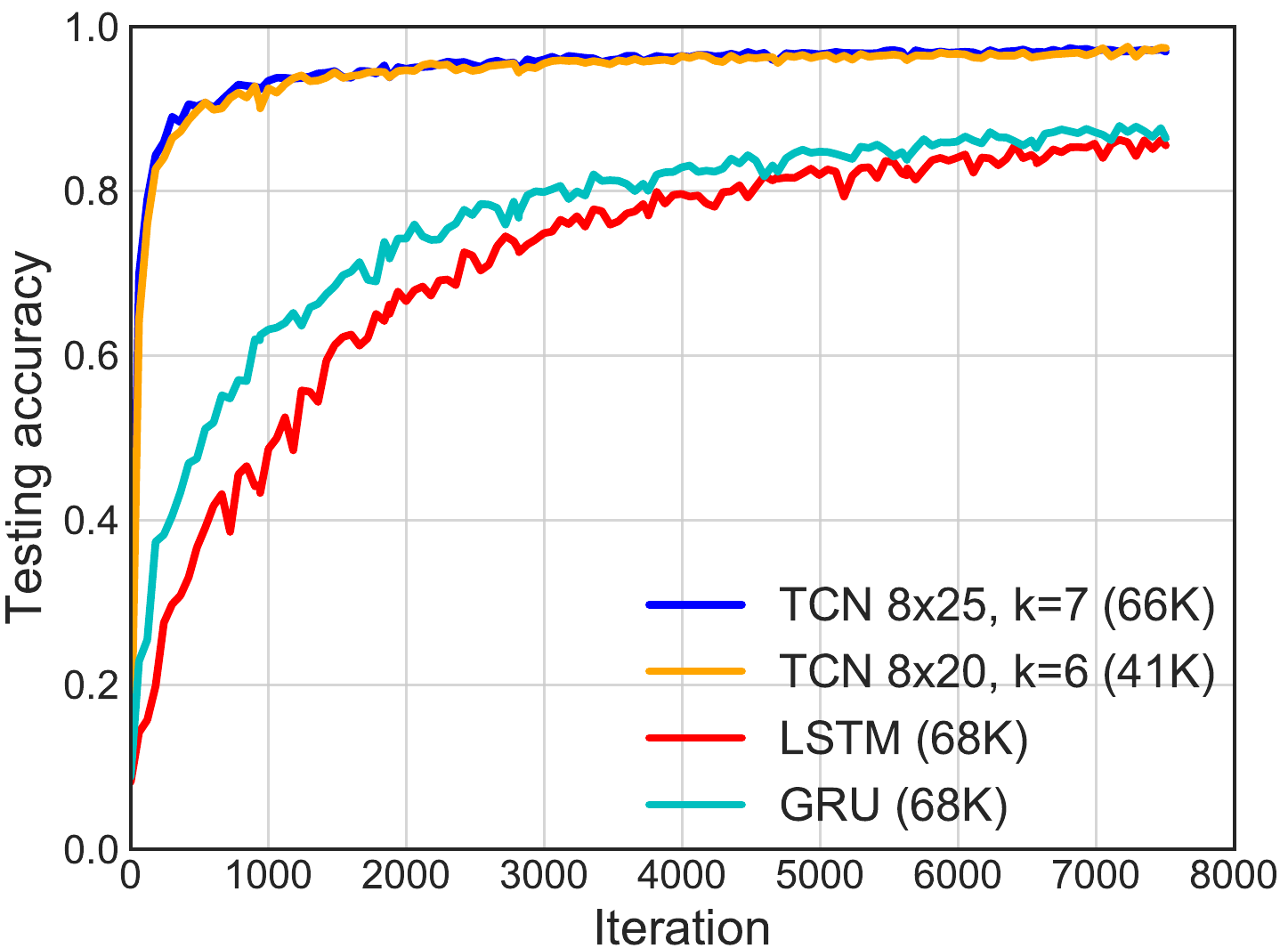}
        \caption{P-MNIST \label{PMNIST-figure}}
    \end{subfigure}
    \vspace{-.05in}
    \caption{Results on Sequential MNIST and P-MNIST. TCNs outperform recurrent architectures.}
\label{seqmnist-figure}
\vspace{-3mm}
\end{figure}

\mypara{Sequential MNIST and P-MNIST.}
Convergence results on sequential and permuted MNIST,
run over 10 epochs, are shown in Figure~\ref{seqmnist-figure}.
All models were configured to have roughly 70K parameters. For both problems, TCNs substantially outperform the recurrent architectures, both in terms of convergence and in final accuracy on the task.  For P-MNIST, TCNs outperform state-of-the-art results (95.9\%) based on recurrent networks with Zoneout and Recurrent BatchNorm~\citep{cooijmans2016recurrent,kruegerZoneout}.

\begin{figure}[t]
    \centering
    \begin{subfigure}[t]{0.23\textwidth}
        \centering
        \includegraphics[width=\textwidth]{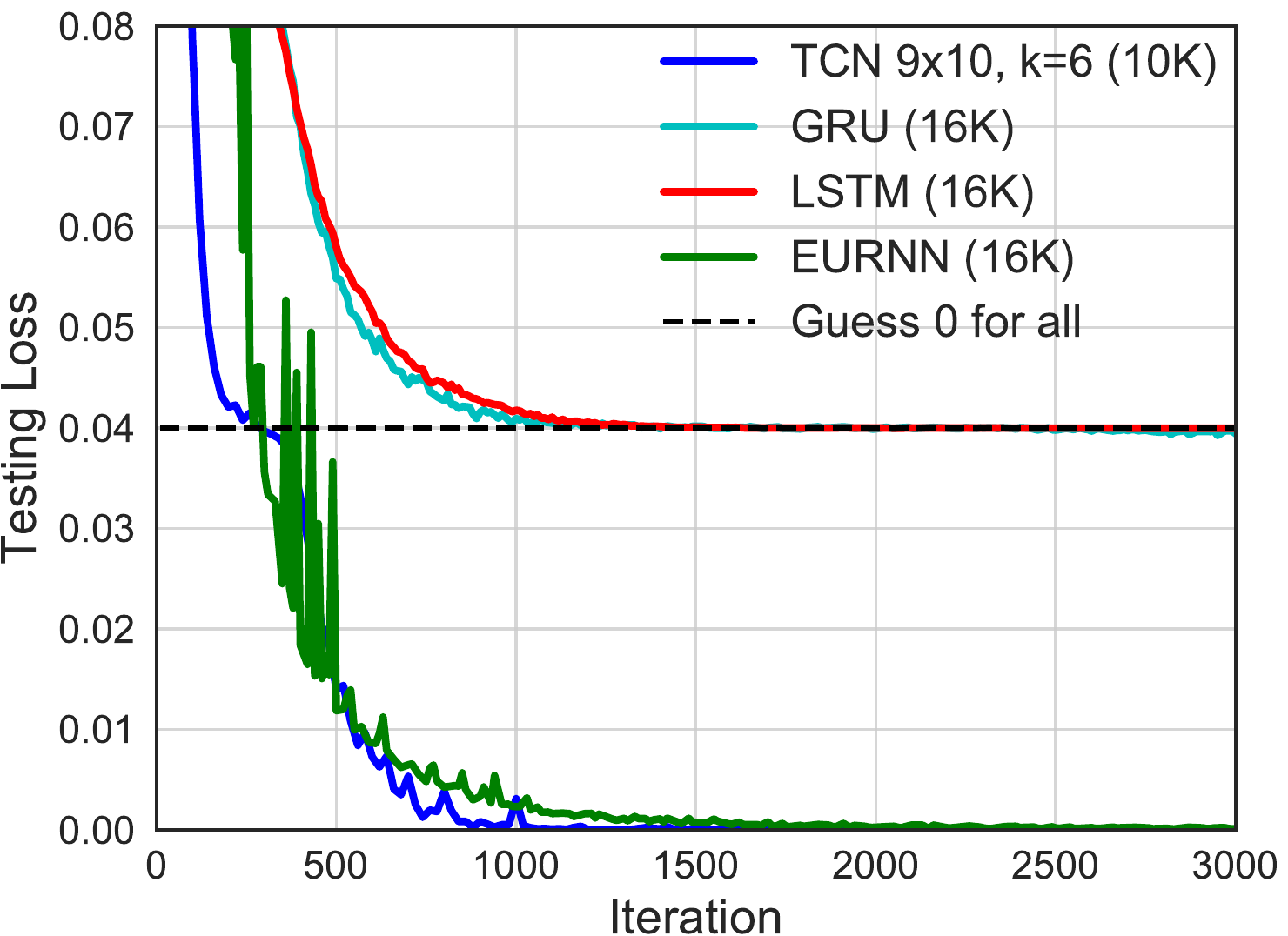}
        \caption{$T=500$ \label{copy-500-figure}}
    \end{subfigure}%
    ~
    \begin{subfigure}[t]{0.23\textwidth}
        \centering
        \includegraphics[width=\textwidth]{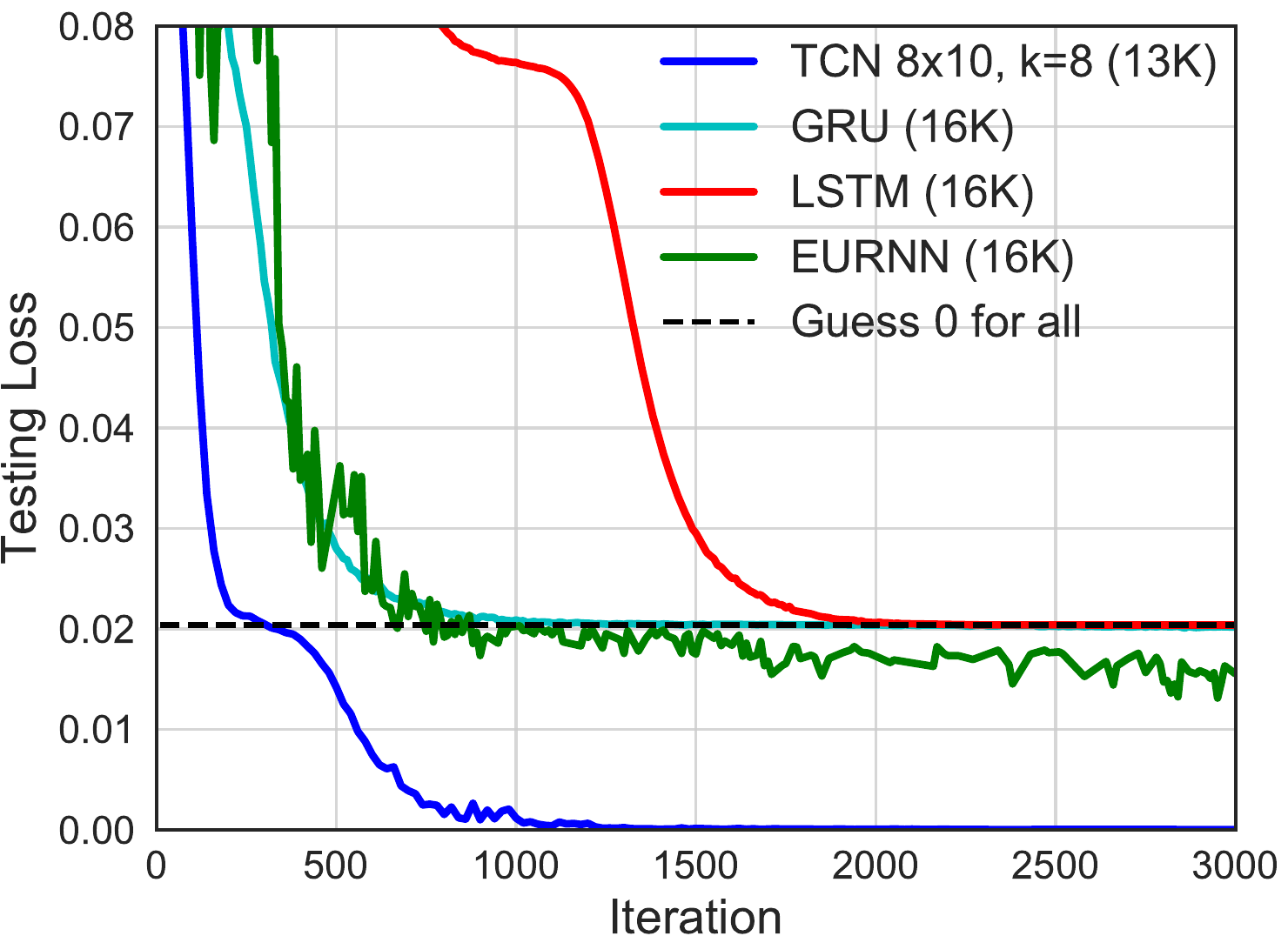}
        \caption{$T=1000$ \label{copy-1000-figure}}
    \end{subfigure}
    \vspace{-.05in}
    \caption{Result on the copy memory task for different sequence lengths $T$. TCNs outperform recurrent architectures.}
    \label{copy-figure}
\vspace{-3mm}
\end{figure}

\mypara{Copy memory.}
Convergence results on the copy memory task are shown in Figure \ref{copy-figure}. TCNs quickly converge to correct answers, while LSTMs and GRUs simply converge to the same loss as predicting all zeros.  In this case we also
compare to the recently-proposed EURNN~\citep{pmlr-v70-jing17a}, which was highlighted to perform well on this task.  While both TCN and EURNN perform well for sequence length ${T=500}$, the TCN has a clear advantage for ${T=1000}$ and longer (in terms of both loss and rate of convergence).

\subsection{Polyphonic Music and Language Modeling}

We now discuss the results on polyphonic music modeling, character-level language modeling, and word-level language modeling. These domains are dominated by recurrent architectures, with many specialized designs developed for these tasks~\cite{Zhang2016NIPS,ha2016hypernetworks,kruegerZoneout,grave2016improving,greffOdyssey,merityRegOpt}.
We mention some of these specialized architectures when useful, but our primary goal is to compare the generic TCN model to similarly generic recurrent architectures, before domain-specific tuning. The results are summarized in Table~\ref{bigtable}.

\mypara{Polyphonic music.}
On Nottingham and JSB Chorales, the
TCN with virtually no tuning outperforms the recurrent models by a
considerable margin, and even outperforms some enhanced recurrent
architectures for this task such as HF-RNN \citep{boulanger2012modeling}
and Diagonal RNN \citep{diagRNN}. Note however that other models such as
the Deep Belief Net LSTM perform better still~\citep{vohra2015modeling};
we believe this is likely due to the fact that the datasets are relatively small, and thus the right regularization method or
generative modeling procedure can improve performance significantly. This is
largely orthogonal to the RNN/TCN distinction, as a similar variant of TCN may
well be possible.

\mypara{Word-level language modeling.}
Language modeling remains one of the primary applications of recurrent
networks and many recent works have focused on optimizing LSTMs for this
task \citep{kruegerZoneout,merityRegOpt}. Our implementation follows
standard practice that ties the weights of encoder and decoder
layers for both TCN and RNNs \citep{press2016using}, which significantly
reduces the number of parameters in the model. For training, we use SGD
and anneal the learning rate by a factor of 0.5 for both TCN and
RNNs when validation accuracy plateaus.

On the smaller PTB corpus, an optimized LSTM architecture (with recurrent
and embedding dropout, etc.) outperforms the TCN, while the TCN outperforms
both GRU and vanilla RNN. However, on the much larger Wikitext-103 corpus and
the LAMBADA dataset~\citep{paperno2016lambada}, without any hyperparameter search, the TCN outperforms the LSTM results of \citet{grave2016improving}, achieving much lower perplexities.

\mypara{Character-level language modeling.}
On character-level language modeling (PTB and text8, accuracy measured in
bits per character), the generic TCN outperforms regularized LSTMs and GRUs
as well as methods such as Norm-stabilized LSTMs \citep{krueger2015regularizing}.
(Specialized architectures exist that outperform all of these, see the
supplement.)

\subsection{Memory Size of TCN and RNNs}
\label{memsize-subsection}

One of the theoretical advantages of recurrent architectures is their unlimited memory: the theoretical ability to retain information through sequences of unlimited length. We now examine specifically how long the different architectures can retain information in practice. We focus on 1) the copy memory task, which is a stress test designed to evaluate long-term, distant information propagation in recurrent networks, and 2) the LAMBADA task, which tests both local and non-local textual understanding.

The copy memory task is perfectly set up to examine a model's ability to retain information for different lengths of time. The requisite retention time can be controlled by varying the sequence length $T$. In contrast to Section \ref{baseline-tasks}, we now focus on the accuracy on the last 10 elements of the output sequence (which are the nontrivial elements that must be recalled). We used models of size 10K for both TCN and RNNs.

\begin{figure}[t]
    \centering
    \includegraphics[width=.4\textwidth]{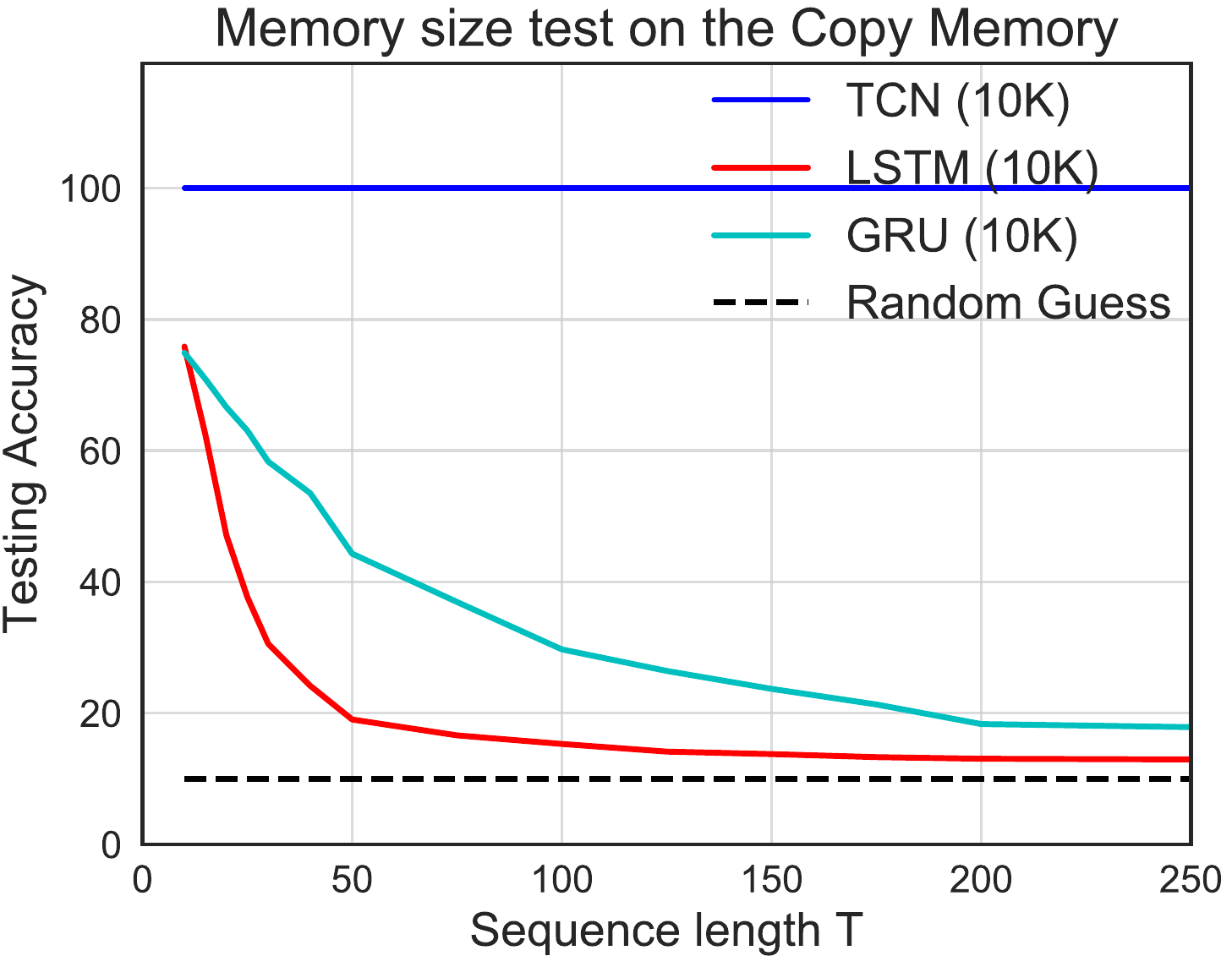}
    \vspace{-2mm}
    \caption{Accuracy on the copy memory task for sequences of different lengths $T$. While TCN exhibits 100\% accuracy for all sequence lengths, the LSTM and GRU degenerate to random guessing as $T$ grows.}
    \label{fig-memory-size}
    \vspace{-3mm}
\end{figure}

The results of this focused study are shown in Figure~\ref{fig-memory-size}. TCNs consistently converge to
100\% accuracy for all sequence lengths, whereas LSTMs and GRUs of the same size quickly degenerate to random guessing as the sequence length $T$ grows. The accuracy of the LSTM falls below 20\% for $T<50$, while the GRU falls below 20\% for $T<200$. These results indicate that TCNs are able to maintain a much longer effective history than their recurrent counterparts.

This observation is backed up on real data by experiments on the large-scale LAMBADA dataset, which is specifically designed to test a model's ability to utilize broad context~\citep{paperno2016lambada}. As shown in Table \ref{bigtable}, TCN outperforms LSTMs and vanilla RNNs by a significant margin in perplexity on LAMBADA, with a substantially smaller network and virtually no tuning. (State-of-the-art results on this dataset are even better, but only with the help of additional memory mechanisms~\cite{grave2016improving}.)

\section{Conclusion \label{conclusion-sec}}

We have presented an empirical evaluation of generic convolutional and recurrent
architectures across a comprehensive suite of sequence modeling tasks.
To this end, we have described a simple temporal convolutional network (TCN) that
combines best practices such as dilations and residual connections with the causal
convolutions needed for autoregressive prediction. The experimental results indicate that TCN models substantially
outperform generic recurrent architectures such as LSTMs and GRUs. We further
studied long-range information propagation in convolutional and recurrent networks,
and showed that the ``infinite memory'' advantage of RNNs is largely absent in
practice. TCNs exhibit longer memory than recurrent architectures
with the same capacity.

Numerous advanced schemes for regularizing and optimizing LSTMs have been proposed \cite{press2016using,kruegerZoneout,merityRegOpt,campos2018skip}. These schemes have significantly advanced the accuracy achieved by LSTM-based architectures on some datasets. The TCN has not yet benefitted from this concerted community-wide investment into architectural and algorithmic elaborations.
We see such investment as desirable and expect it to yield advances in TCN performance that are commensurate with the advances seen in recent years in LSTM performance. We will release the code for our project to encourage this exploration.

The preeminence enjoyed by recurrent networks in sequence modeling may
be largely a vestige of history. Until recently, before the introduction of architectural
elements such as dilated convolutions and residual connections, convolutional
architectures were indeed weaker. Our results indicate that with these elements,
a simple convolutional architecture is more effective across diverse sequence
modeling tasks than recurrent architectures such as LSTMs. Due to the comparable
clarity and simplicity of TCNs, we conclude that convolutional networks should
be regarded as a natural starting point and a powerful toolkit for sequence modeling.

\setlength{\bibsep}{1.25mm}
{\small \bibliography{tcn}}
\bibliographystyle{icml2018}

\appendix

\twocolumn[
  \icmltitle{An Empirical Evaluation of Generic Convolutional and Recurrent
  Networks for Sequence Modeling \\ \emph{Supplementary Material}}
  \vspace{5mm}
]

\section{Hyperparameters Settings \label{app-params}}

\subsection{Hyperparameters for TCN}

\begin{table*}[!t]
\def\arraystretch{1.3}
\small
\centering
\caption{TCN parameter settings for experiments in Section \ref{experiments-sec}.
\label{param-table}}
{\begin{tabular}{ |c|c|c|c|c|c|c|c| }
\hline
\multicolumn{8}{ |c| }{\textsc{TCN Settings}} \\
\hline
\textbf{Dataset/Task} & \textbf{Subtask} & $k$ & $n$ & \textbf{Hidden} & \textbf{Dropout} & \textbf{Grad Clip} & \textbf{Note}   \\ \hline
\multirow{3}{*}{The Adding Problem}  & $T=200$ & 6 & 7 & 27 & \multirow{3}{*}{0.0} & \multirow{3}{*}{N/A} & \\
                                     & $T=400$ & 7 & 7 & 27 &  &  & \\
                                     & $T=600$ & 8 & 8 & 24 &  &  & \\
\hline
\multirow{2}{*}{Seq. MNIST}    & \multirow{2}{*}{-} & 7 & 8 & 25 & \multirow{2}{*}{0.0} & \multirow{2}{*}{N/A} & \\
                                     &          & 6 & 8 & 20 &  &  & \\
\hline
\multirow{2}{*}{Permuted MNIST}      & \multirow{2}{*}{-} & 7 & 8 & 25 & \multirow{2}{*}{0.0} & \multirow{2}{*}{N/A} & \\
                                     &          & 6 & 8 & 20 &  &  & \\
\hline
\multirow{3}{*}{Copy Memory Task}    & $T=500$  & 6 & 9 & 10 & \multirow{3}{*}{0.05} & \multirow{3}{*}{1.0} & \multirow{3}{*}{RMSprop 5e-4} \\
                                     & $T=1000$ & 8 & 8 & 10 &  &  & \\
                                     & $T=2000$ & 8 & 9 & 10 &  &  & \\
\hline
Music JSB Chorales                   & -        & 3 & 2 & 150 & 0.5 & 0.4 & \\
\hline
Music Nottingham                     & -        & 6 & 4 & 150 & 0.2 & 0.4 & \\
\hline
\multirow{3}{*}{Word-level LM} & PTB & 3 & 4 & 600 & 0.5 & \multirow{3}{*}{0.4} & Embed. size 600 \\
                               & Wiki-103 & 3 & 5 & 1000 & \multirow{2}{*}{0.4} &  & Embed. size 400 \\
                               & LAMBADA & 4 & 5 & 500 &  &  & Embed. size 500 \\
\hline
\multirow{2}{*}{Char-level LM} & PTB & 3 & 3 & 450 & \multirow{2}{*}{0.1} & \multirow{2}{*}{0.15} & \multirow{2}{*}{\centering Embed. size 100} \\
                               & text8   & 2 & 5 & 520 &  &  &  \\
\hline
\end{tabular}}
\end{table*}

Table \ref{param-table} lists the hyperparameters we used when applying the generic TCN model on various tasks and datasets. The most important factor for picking parameters is to make sure
that the TCN has a sufficiently large receptive field by choosing $k$
and $d$ that can cover the amount of context needed for the task.

As discussed in Section \ref{experiments-sec}, the number
of hidden units was chosen so that the model size is
approximately at the same level as the recurrent models with which we are comparing.
In Table \ref{param-table}, a gradient clip of N/A means no gradient clipping was
applied. In larger tasks (e.g., language modeling), we
empirically found that gradient clipping (we randomly picked a threshold from $[0.3, 1]$)
helps with regularizing TCN and accelerating convergence.

All weights were initialized from a Gaussian disitribution $\mathcal{N}(0, 0.01)$. In general, we found TCN to be relatively insensitive to hyperparameter changes, as long as the effective history (i.e., receptive field) size is sufficient.

\subsection{Hyperparameters for LSTM/GRU}

Table \ref{lstm-param-table} reports hyperparameter settings that were used for the LSTM.
These values are picked from hyperparameter search for LSTMs that have up to 3
layers, and the optimizers are chosen from \{SGD, Adam, RMSprop, Adagrad\}. For
certain larger datasets, we adopted the settings used in prior work (e.g., \citet{grave2016improving} on Wikitext-103). GRU hyperparameters were chosen in a similar fashion, but typically with more hidden units than in LSTM to keep the total
network size approximately the same (since a GRU cell is more compact).

\section{State-of-the-Art Results}

As previously noted, the generic TCN and LSTM/GRU models we used can be outperformed by more specialized architectures on some tasks. State-of-the-art results are summarized in Table \ref{sota-table}. The same TCN architecture is used across all tasks. Note that the size of the state-of-the-art model may be different from the size of the TCN.

\section{Effect of Filter Size and Residual Block}
\label{app-control}

In this section we briefly study the effects of different components of a TCN
layer. Overall, we believe dilation is required
for modeling long-term dependencies, and so we mainly focus on two other
factors here: the filter size $k$ used by each layer, and the effect of
residual blocks.

\begin{figure*}[!t]
    \centering
    \begin{subfigure}[t]{0.32\textwidth}
        \centering
        \includegraphics[width=\textwidth]{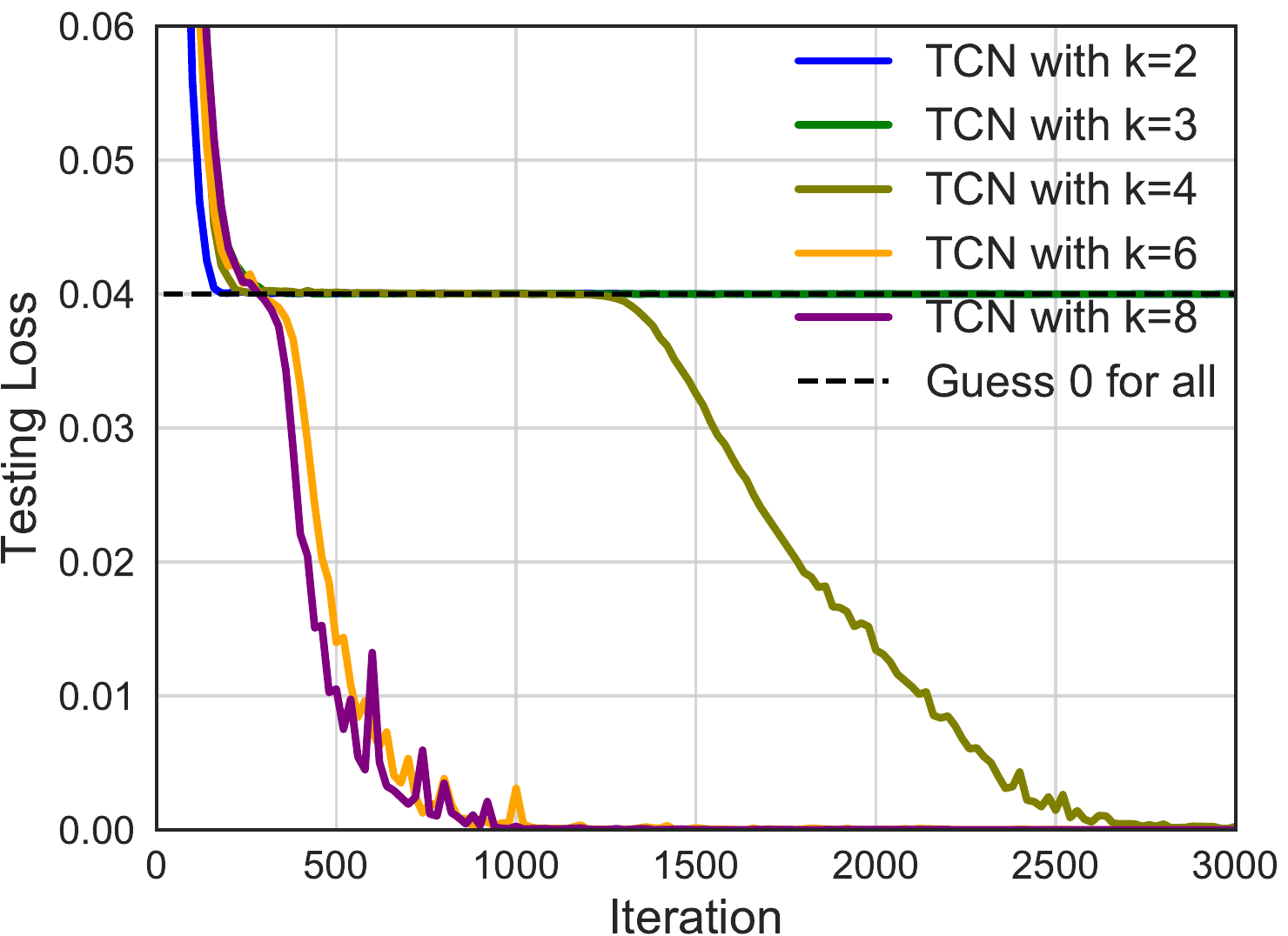}
        \caption{Different $k$ on Copy Memory Task \label{control-k-copy-figure}}
    \end{subfigure}%
    ~
    \begin{subfigure}[t]{0.32\textwidth}
        \centering
        \includegraphics[width=\textwidth]{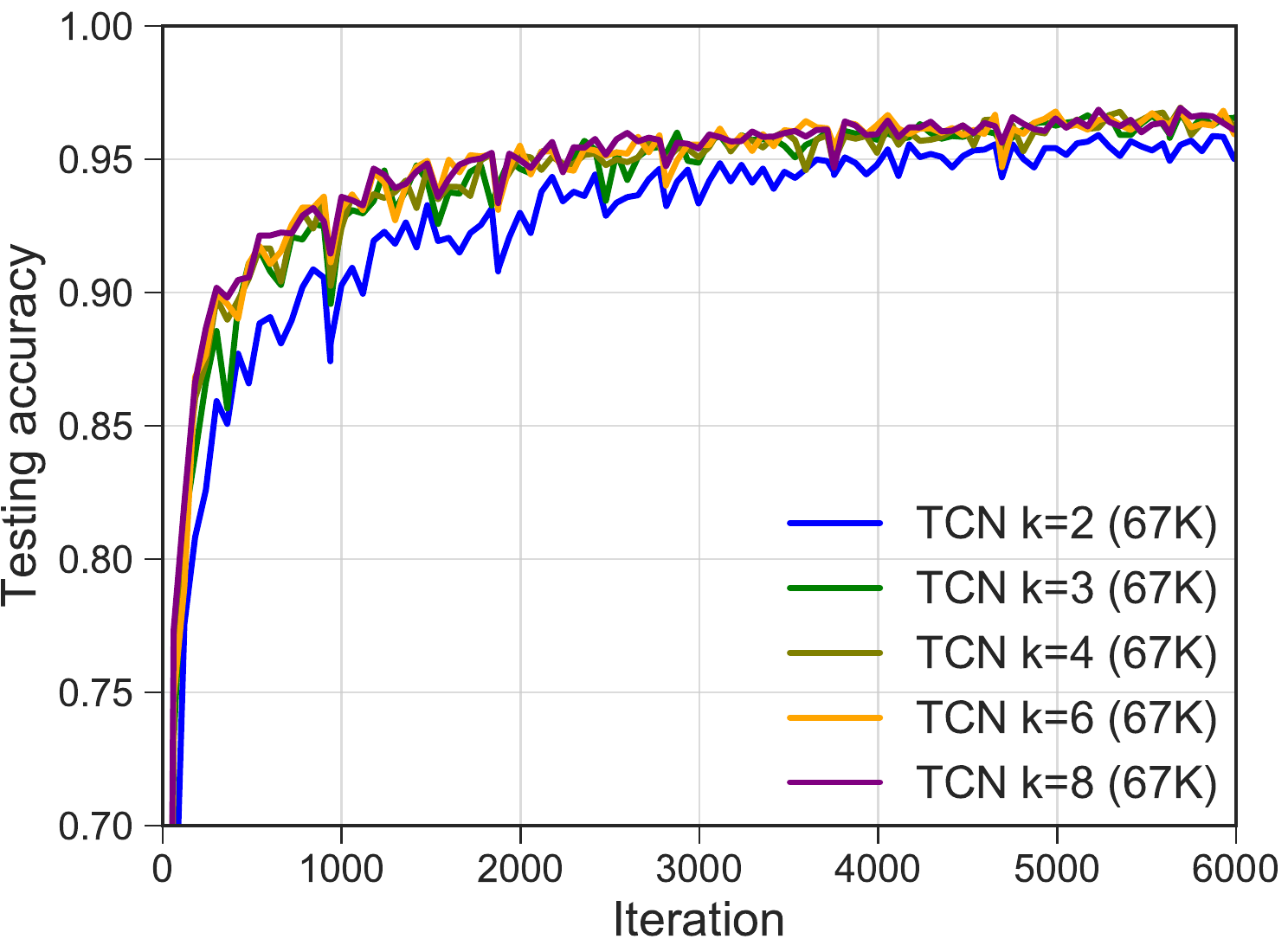}
        \caption{Different $k$ on P-MNIST \label{control-k-pmnist-figure}}
    \end{subfigure}
    ~
    \begin{subfigure}[t]{0.32\textwidth}
        \centering
        \includegraphics[width=\textwidth]{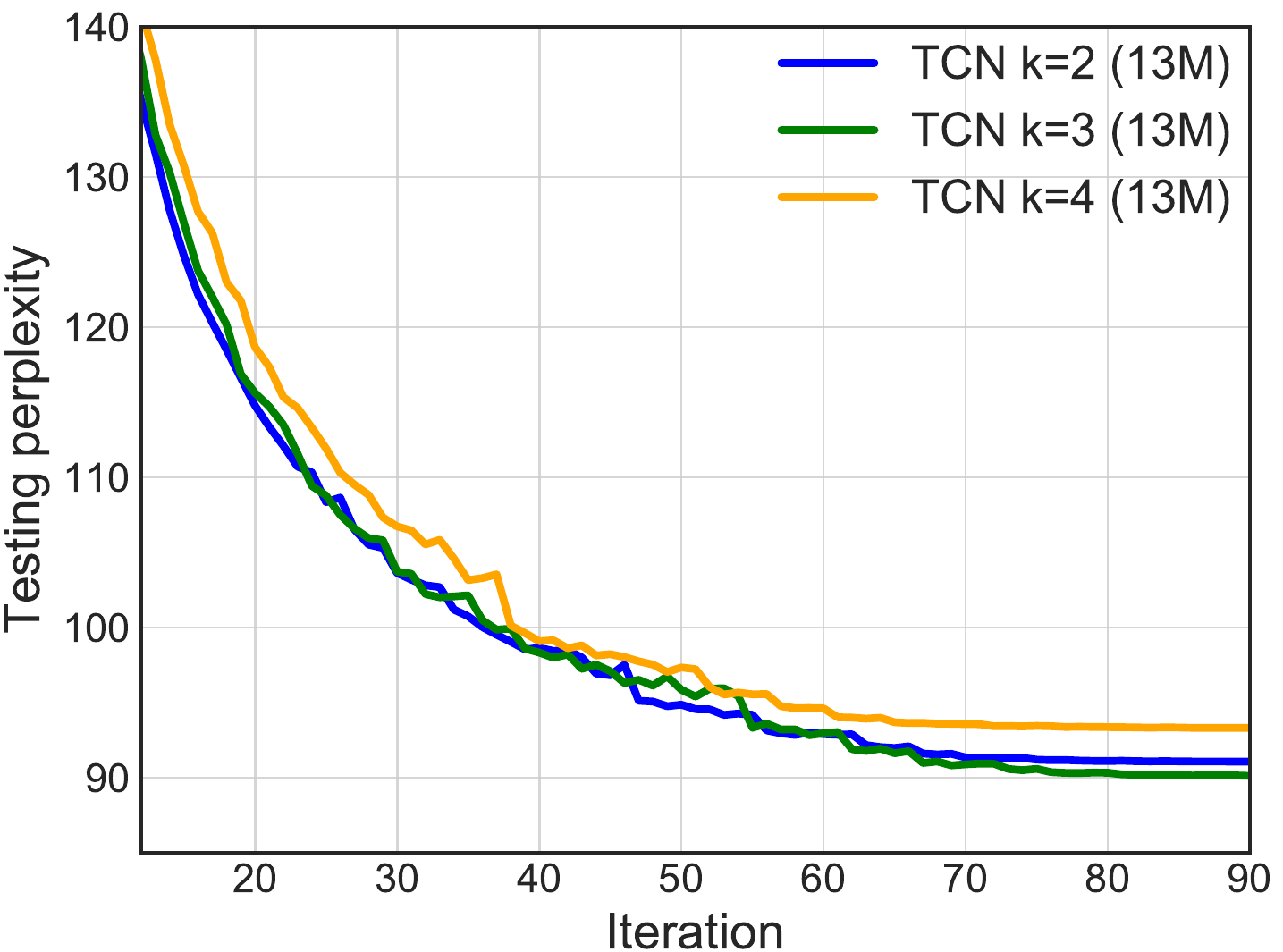}
        \caption{Different $k$ on PTB (word) \label{control-k-ptb-figure}}
    \end{subfigure}
    ~
    \begin{subfigure}[t]{0.323\textwidth}
        \centering
        \includegraphics[width=\textwidth]{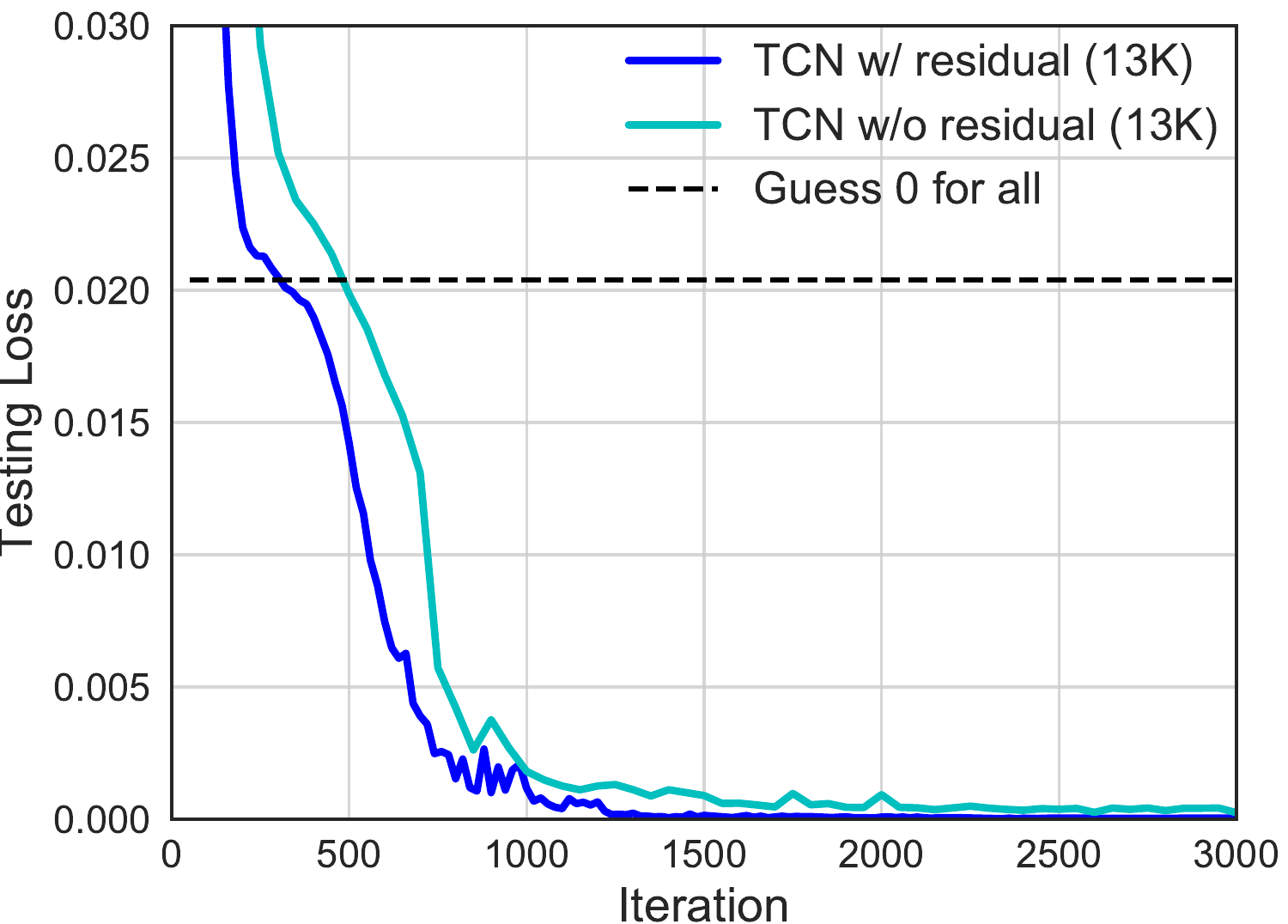}
        \caption{Residual on Copy Memory Task \label{control-res-copy-figure}}
    \end{subfigure}%
    ~
    \begin{subfigure}[t]{0.32\textwidth}
        \centering
        \includegraphics[width=\textwidth]{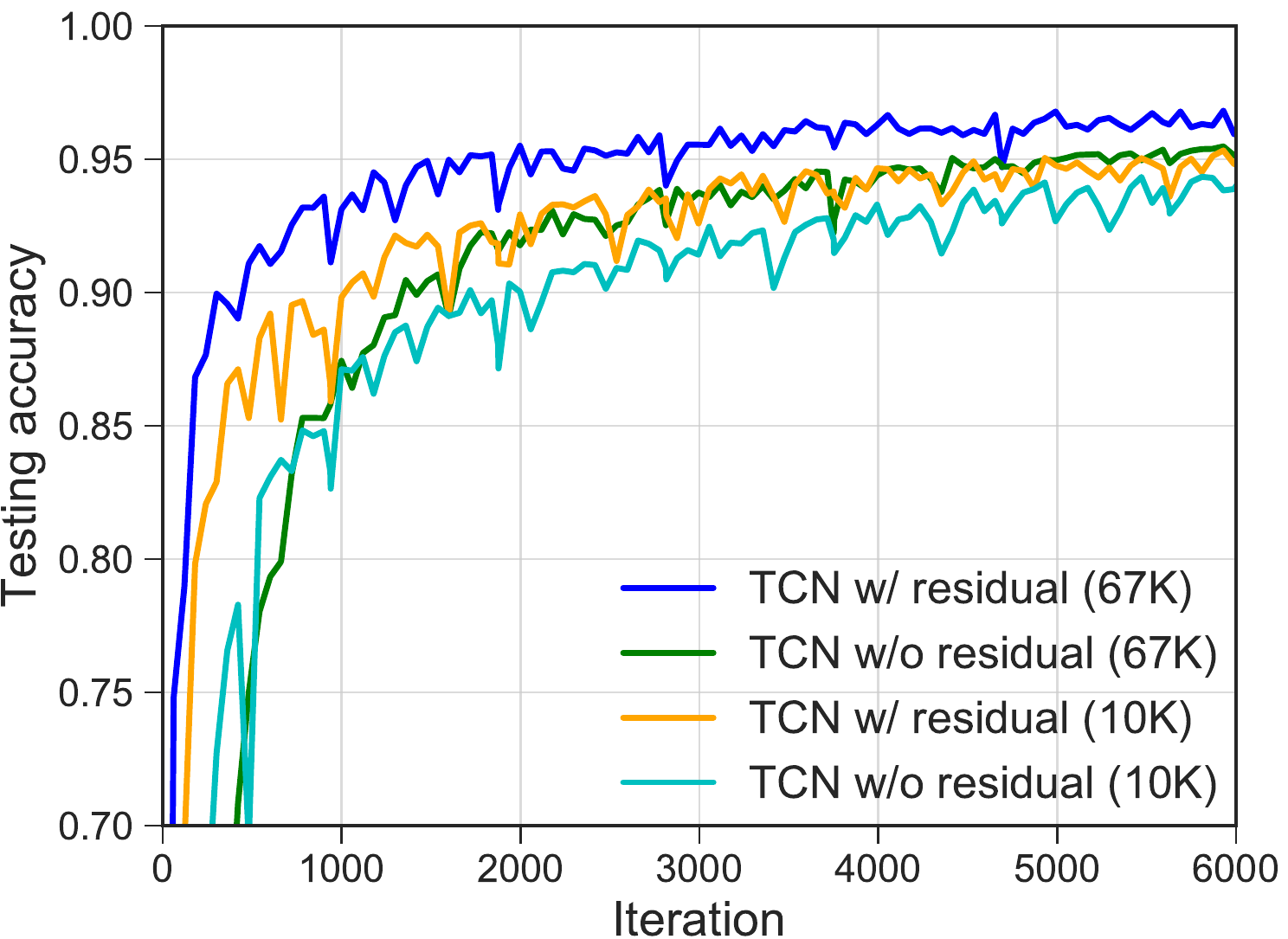}
        \caption{Residual on P-MNIST \label{control-res-pmnist-figure}}
    \end{subfigure}
    ~
    \begin{subfigure}[t]{0.316\textwidth}
        \centering
        \includegraphics[width=\textwidth]{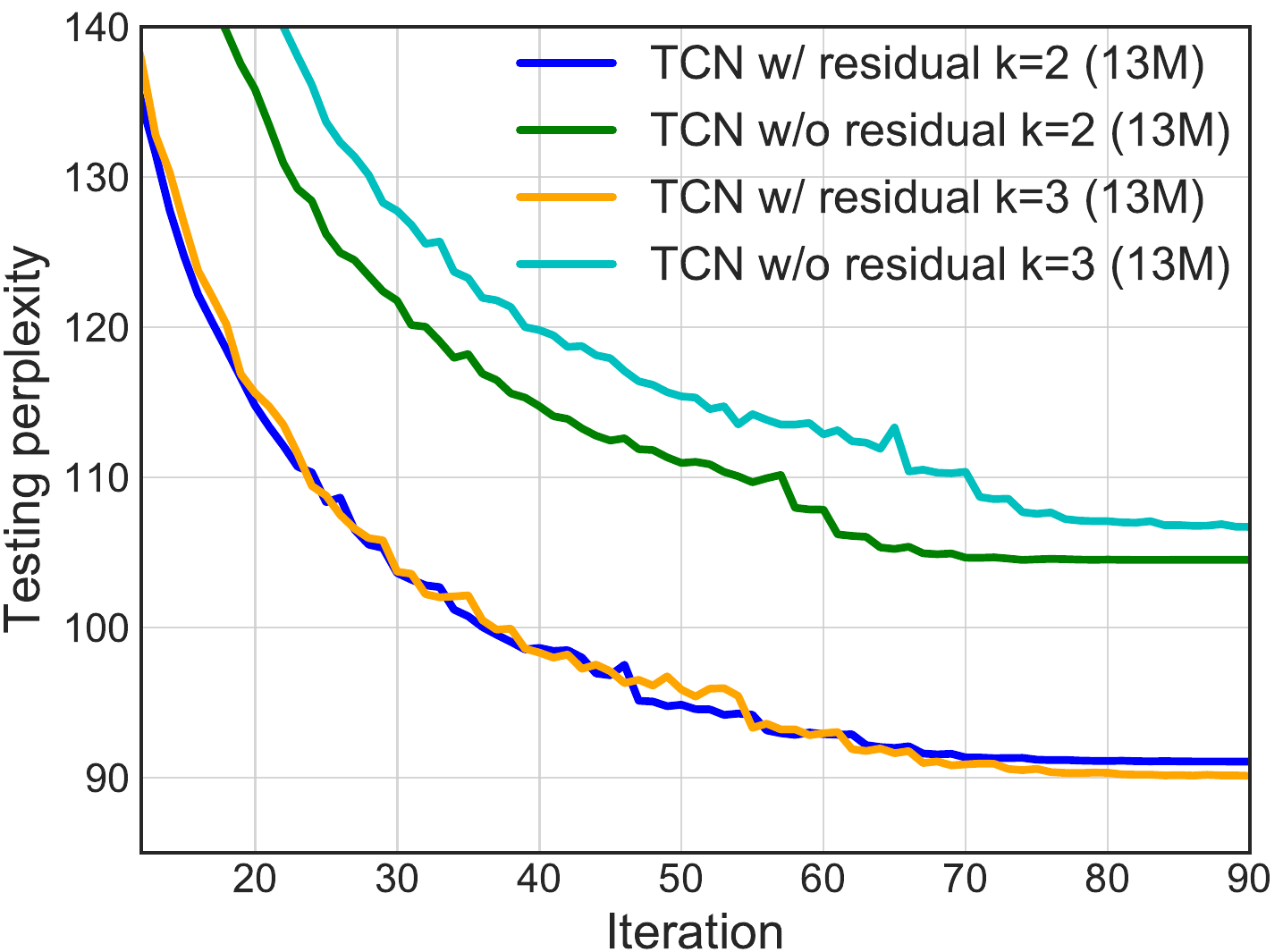}
        \caption{Residual on PTB (word) \label{control-res-ptb-figure}}
    \end{subfigure}
    \caption{Controlled experiments that study the effect of different components of
    the TCN model.
    \label{control-figure}}
\end{figure*}

We perform a series of controlled experiments, with the results of the ablative
analysis shown in Figure \ref{control-figure}. As before, we kept the model size
and depth exactly the same for different models, so that the dilation
factor is strictly controlled. The experiments were conducted on three
different tasks: copy memory, permuted MNIST (P-MNIST), and Penn Treebank word-level language modeling. These experiments confirm that both factors (filter size and residual connections) contribute to sequence modeling performance.

\textbf{Filter size $k$.} In both the copy memory and the P-MNIST tasks,
we observed faster convergence and better accuracy for larger filter sizes. In particular,
looking at Figure \ref{control-k-copy-figure}, a TCN with filter size $\leq 3$ only
converges to the same level as random guessing. In contrast, on word-level language
modeling, a smaller kernel with filter size of $k=3$ works best. We
believe this is because a smaller kernel (along with fixed dilation) tends to focus
more on the local context, which is especially important for PTB language modeling (in
fact, the very success of $n$-gram models suggests that only a relatively short memory
is needed for modeling language).

\textbf{Residual block.} In all three scenarios that we compared here, we observed
that the residual function stabilized training and brought faster convergence
with better final results. Especially in language modeling, we found
that residual connections contribute substantially to performance (See Figure
\ref{control-res-ptb-figure}).

\section{Gating Mechanisms}
\label{gating-mech}

One component that had been used in prior work on convolutional
architectures for language modeling is the gated activation \cite{waveNet,dauphinGatedConv}. We have
chosen not to use gating in the generic TCN model. We now examine this choice more closely.

\citet{dauphinGatedConv} compared the effects of gated linear units (GLU) and gated tanh units (GTU),
and adopted GLU in their non-dilated gated ConvNet. Following the same choice, we now compare TCNs using ReLU and TCNs with gating (GLU), represented by an elementwise product between two convolutional layers, with one of them
also passing through a sigmoid function $\sigma(x)$. Note that the gates architecture uses approximately twice as many convolutional layers as the  ReLU-TCN.

The results are shown in Table~\ref{gated-table}, where we kept the number of model
parameters at about the same size. The GLU does further
improve TCN accuracy on certain language modeling datasets like PTB, which agrees
with prior work. However, we do not observe comparable benefits on other tasks, such
as polyphonic music modeling or synthetic stress tests that require
longer information retention. On the copy memory task with $T=1000$, we found
that TCN with gating converged to a worse result than
TCN with ReLU (though still better than recurrent models).

\newpage

\begin{table*}[t!]
\def\arraystretch{1.3}
\small
\centering
\caption{LSTM parameter settings for experiments in Section \ref{experiments-sec}.\label{lstm-param-table}}
{\begin{tabular}{ |c|c|c|c|c|c|c|c| }
\hline
\multicolumn{8}{ |c| }{\textsc{LSTM Settings (Key Parameters)}} \\
\hline
\textbf{Dataset/Task} & \textbf{Subtask} & $n$ & \textbf{Hidden} & \textbf{Dropout} & \textbf{Grad Clip} & \textbf{Bias} & \textbf{Note}   \\ \hline
\multirow{3}{*}{The Adding Problem}  & $T=200$ & 2 & 77 & \multirow{3}{*}{0.0} & 50 & 5.0 & SGD 1e-3 \\
                                     & $T=400$ & 2 & 77 &   & 50 & 10.0 & Adam 2e-3 \\
                                     & $T=600$ & 1 & 130 &  &  5 & 1.0 & - \\
\hline
Seq. MNIST    & - & 1 & 130 & 0.0 & 1 & 1.0 & RMSprop 1e-3 \\
\hline
Permuted MNIST      & - & 1 & 130 & 0.0 & 1 & 10.0 & RMSprop 1e-3 \\
\hline
\multirow{3}{*}{Copy Memory Task}    & $T=500$  & 1 & 50 & \multirow{3}{*}{0.05} & 0.25 & \multirow{3}{*}{-} & \multirow{3}{*}{RMSprop/Adam} \\
                                     & $T=1000$ & 1 & 50 &  &  1  &  & \\
                                     & $T=2000$ & 3 & 28 &  &  1  &  & \\
\hline
Music JSB Chorales                   & -  & 2 & 200 & 0.2 & 1 & 10.0 & SGD/Adam \\
\hline
\multirow{2}{*}{Music Nottingham}    & \multirow{2}{*}{-}  & 3 & 280 & \multirow{2}{*}{0.1} & 0.5 & - & \multirow{2}{*}{Adam 4e-3} \\
                                     &                     & 1 & 500 &  & 1 & - & \\
\hline
\multirow{3}{*}{Word-level LM} & PTB & 3 & 700 & 0.4 & 0.3 & 1.0 & SGD 30, Emb. 700, etc. \\
                               & Wiki-103 & - & - & - & - & - & \citet{grave2016improving} \\
                               & LAMBADA & - & - & - & - & - & \citet{grave2016improving} \\
\hline
\multirow{2}{*}{Char-level LM} & PTB & 2 & 600 & 0.1 & 0.5 & - & Emb. size 120 \\
                               & text8 & 1 & 1024 & 0.15 & 0.5 & - & Adam 1e-2 \\
\hline
\end{tabular}}
\end{table*}

\begin{table*}[ht!]
\def\arraystretch{1.3}
\small
\centering
\caption{State-of-the-art (SoTA) results for tasks in Section \ref{experiments-sec}.
\label{sota-table}}
{\begin{tabular}{ |c||c|c||c|c|c| }
\hline
\multicolumn{6}{ |c| }{\textsc{TCN vs. SoTA Results}} \\
\hline
\textbf{Task} & \textbf{TCN Result} & \textbf{Size} & \textbf{SoTA} & \textbf{Size} & \textbf{Model}  \\ \hline
Seq. MNIST (acc.) & 99.0 & 21K & 99.0 & 21K & Dilated GRU \citep{chang2017dilated} \\ \hline
P-MNIST (acc.) & 97.2 & 42K & 95.9 & 42K & Zoneout \citep{kruegerZoneout} \\ \hline
Adding Prob. 600 (loss) & 5.8e-5 & 70K & 5.3e-5 & 70K & Regularized GRU \\ \hline
Copy Memory 1000 (loss) & 3.5e-5 & 70K & 0.011 & 70K & EURNN \citep{pmlr-v70-jing17a} \\ \hline
JSB Chorales (loss) & 8.10 & 300K & 3.47 & - & DBN+LSTM \citep{vohra2015modeling} \\ \hline
Nottingham (loss) & 3.07 & 1M & 1.32 & - & DBN+LSTM \citep{vohra2015modeling} \\ \hline
\multirow{2}{*}{Word PTB (ppl)} & \multirow{2}{*}{\wordptbres} & \multirow{2}{*}{13M} & \multirow{2}{*}{47.7} & \multirow{2}{*}{22M} & \multirow{2}{4cm}{\centering AWD-LSTM-MoS + Dynamic Eval. \citep{yang2018breaking}} \\
& & & & & \\
\hline
\multirow{2}{*}{Word Wiki-103 (ppl)} & \multirow{2}{*}{\wordwikires} & \multirow{2}{*}{148M} & \multirow{2}{*}{40.4} & \multirow{2}{*}{$>$300M} & \multirow{2}{4cm}{\centering Neural Cache Model (Large) \citep{grave2016improving}} \\
& & & & & \\
\hline
\multirow{2}{*}{Word LAMBADA (ppl)} & \multirow{2}{*}{\wordlambadares} & \multirow{2}{*}{56M} & \multirow{2}{*}{138} & \multirow{2}{*}{$>$100M} & \multirow{2}{4cm}{\centering Neural Cache Model (Large) \citep{grave2016improving}} \\
& & & & & \\
 \hline
\multirow{2}{*}{Char PTB (bpc)} & \multirow{2}{*}{\charptbres} & \multirow{2}{*}{3M} & \multirow{2}{*}{1.22} & \multirow{2}{*}{14M} & \multirow{2}{3.8cm}{\centering 2-LayerNorm HyperLSTM \citep{ha2016hypernetworks}} \\
& & & & & \\
\hline
Char text8 (bpc) & \chartextres & 4.6M & 1.29 & $>$12M & HM-LSTM \citep{chung2016hierarchical} \\ \hline
\end{tabular}}
\end{table*}

\begin{table*}[t!]
\def\arraystretch{1.4}
\small
\centering
\caption{An evaluation of gating in TCN. A plain TCN is compared to a TCN that uses gated activations.
\vspace{.1in}
\label{gated-table}}
{\begin{tabular}{ |c||c|c| }
\hline
\textbf{Task} & \textbf{TCN} & \textbf{TCN + Gating} \\ \hline
Sequential MNIST (acc.)       & \textbf{99.0}      & \textbf{99.0} \\ \hline
Permuted MNIST (acc.)          & \textbf{97.2}    & 96.9 \\ \hline
Adding Problem $T=600$ (loss) & \textbf{5.8e-5}    & \textbf{5.6e-5} \\ \hline
Copy Memory $T=1000$ (loss) & \textbf{3.5e-5}    & 0.00508 \\ \hline
JSB Chorales (loss)     & \textbf{8.10}      & 8.13\\ \hline
Nottingham (loss)       & \textbf{3.07}      & 3.12 \\ \hline
\multirow{2}{*}{Word-level PTB (ppl)} & \multirow{2}{*}{\wordptbres} & \textbf{\multirow{2}{*}{87.94}} \\
& & \\
\hline
\multirow{2}{*}{Char-level PTB (bpc)} & \multirow{2}{*}{\charptbres} & \textbf{\multirow{2}{*}{1.306}} \\
& & \\
\hline
Char text8 (bpc) & \textbf{\chartextres} & 1.485 \\ \hline
\end{tabular}}
\end{table*}

%%%%%%%%%%%%%%%%%%%%%%%%%%%%%%%%%%%%%%%%%%%%%%%%%%%%%%%%%%%%%%%%%%%%%%%%%%%%%%%
%%%%%%%%%%%%%%%%%%%%%%%%%%%%%%%%%%%%%%%%%%%%%%%%%%%%%%%%%%%%%%%%%%%%%%%%%%%%%%%

\end{document}